\documentclass[10pt,twocolumn,letterpaper]{article}

\usepackage[pagenumbers]{cvpr} 
\usepackage{cvpr}
\usepackage{times}
\usepackage{epsfig}
\usepackage{graphicx}
\usepackage{amsmath}
\usepackage{amssymb}
\usepackage{epsfig}
\usepackage{graphicx}
\usepackage{amsmath}
\usepackage{amssymb}
\usepackage{todonotes}
\usepackage{lineno}
\usepackage{colortbl}
\usepackage{caption}
\usepackage{subcaption}
\usepackage{booktabs}
\usepackage{xcolor}
\usepackage{soul}
\usepackage{xr}
\usepackage{multirow}
\usepackage{algorithm}
\usepackage{algpseudocode}
\usepackage[accsupp]{axessibility}

\newcommand\red[1]{\textcolor{red}{#1}}

\definecolor{darkgreen}{rgb}{0.0, 0.4, 0.0}
\definecolor{orange}{rgb}{1.0, 0.49, 0.0}
\definecolor{purple}{rgb}{0.54, 0.17, 0.89}
\definecolor{ForestGreen}{RGB}{34,139,34}
\definecolor{ModernBlue}{RGB}{0,153,153}
\definecolor{LightModernBlue}{RGB}{0,204,204}
\definecolor{DarkPink}{RGB}{204,0,102}

\newcommand{\datagenerator}{\texttt{M$^{3}$Act}}

\newcommand\blfootnote[1]{
    \begingroup
    \renewcommand\thefootnote{}\footnote{#1}
    \addtocounter{footnote}{-1}
    \endgroup
}

\definecolor{cvprblue}{rgb}{0.21,0.49,0.74}
\usepackage[pagebackref,breaklinks,colorlinks,citecolor=cvprblue]{hyperref}

\def\paperID{2282} 
\def\confName{CVPR}
\def\confYear{2024}

\title{{\datagenerator}: Learning from Synthetic Human Group Activities}

\author{Che-Jui Chang\textsuperscript{\textdagger}\\
Rutgers University\\
{\tt\small chejui.chang@rutgers.edu}
\and
Danrui Li\\ 
Rutgers University\\
{\tt\small danrui.li@rutgers.edu}
\and
Deep Patel\\ 
NEC Laboratories\\
{\tt\small dpatel@nec-labs.com}
\and
Parth Goel\\ 
Rutgers University\\
{\tt\small goel.parth210@gmail.com}
\and
Honglu Zhou\\ 
NEC Laboratories\\
{\tt\small hozhou@nec-labs.com}
\and
Seonghyeon Moon\textsuperscript{\textdagger}\\ 
Rutgers University\\
{\tt\small sm206@cs.rutgers.edu}
\and
Samuel S. Sohn\\ 
Rutgers University\\
{\tt\small samuel.sohn@rutgers.edu}
\and
Sejong Yoon\\
The College of New Jersey\\
{\tt\small yoons@tcnj.edu}
\and
Vladimir Pavlovic\\
Rutgers University\\
{\tt\small vladimir@rutgers.edu}
\and
Mubbasir Kapadia\\
Roblox\\
{\tt\small mkapadia@roblox.com}
}

\begin{document}

\twocolumn[{%
\renewcommand\twocolumn[1][]{#1}%
\maketitle
\vspace{-20pt}


}]



\begin{abstract}
\vspace{-7.5pt}
The study of complex human interactions and group activities has become a focal point in human-centric computer vision. 
However, progress in related tasks is often hindered by the challenges of obtaining large-scale labeled datasets from real-world scenarios.
To address the limitation, we introduce {\datagenerator}, a synthetic data generator for \textbf{m}ulti-view \textbf{m}ulti-group \textbf{m}ulti-person human atomic \textbf{act}ions and group \textbf{act}ivities. 
Powered by Unity Engine, {\datagenerator} features multiple semantic groups, highly diverse and photorealistic images, and a comprehensive set of annotations, which facilitates the learning of human-centered tasks across single-person, multi-person, and multi-group conditions.
We demonstrate the advantages of {\datagenerator} across three core experiments.
The results suggest our synthetic dataset can significantly improve the performance of several downstream methods and replace real-world datasets to reduce cost.
Notably, {\datagenerator} improves the state-of-the-art MOTRv2 on DanceTrack dataset, leading to a hop on the leaderboard from $10^{th}$ to $2^{nd}$ place. 
Moreover, {\datagenerator} opens new research for controllable 3D group activity generation. 
We define multiple metrics and propose a competitive baseline for the novel task. 
Our code and data are available at our project page: \href{http://cjerry1243.github.io/M3Act}{http://cjerry1243.github.io/M3Act}.
\blfootnote{\textsuperscript{\textdagger}Work done during internship at Roblox}

\end{abstract}
\vspace{-15pt}

\begin{figure*}[t]
    \centering
    \includegraphics[width=\linewidth]{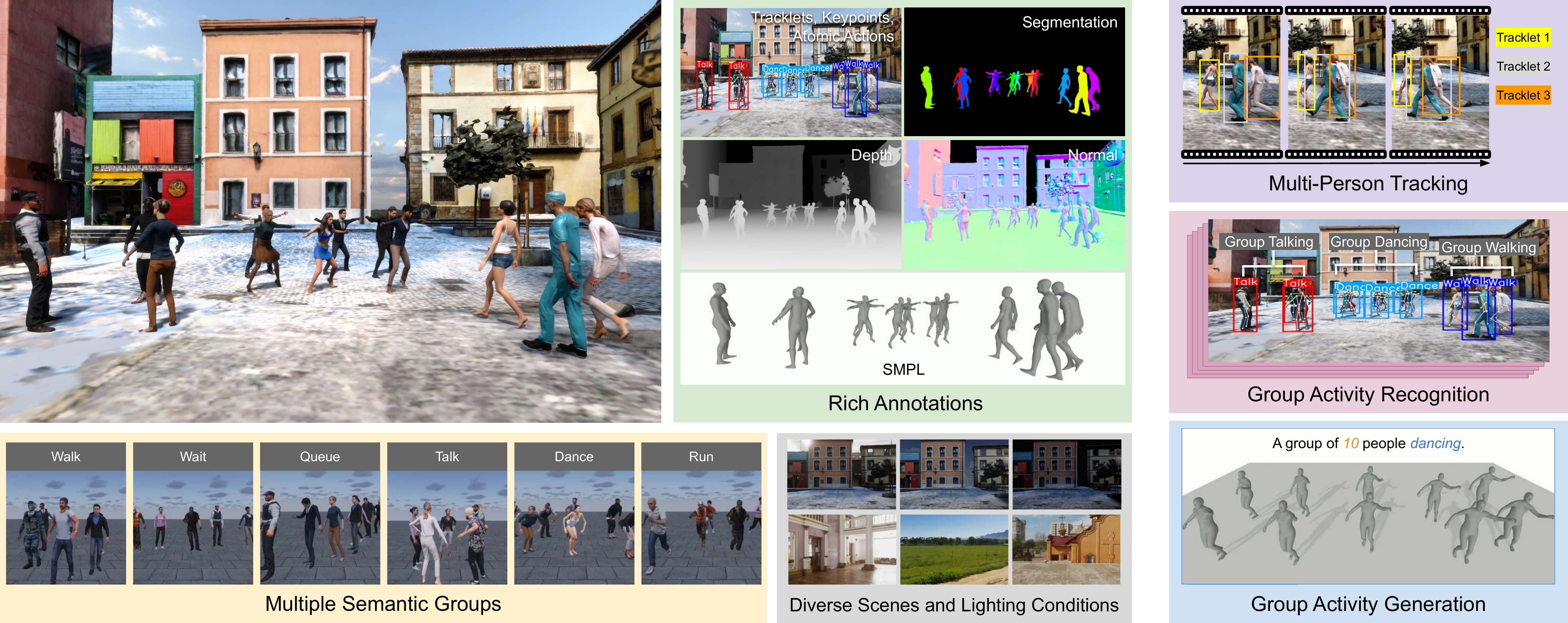}
    \vspace{-15pt}
    \captionof{figure}{\textbf{{\datagenerator} is a large-scale synthetic data generator designed to support multi-person and multi-group research topics.} {\datagenerator} features multiple semantic groups and produces highly diverse and photorealistic videos with a rich set of annotations suitable for human-centered tasks including multi-person tracking, group activity recognition, and controllable human group activity generation.}
\label{fig:teaser}
\vspace{-12pt}
\end{figure*}%

\section{Introduction}
\vspace{-5pt}
Understanding \textit{collective human activities} and \textit{social groups} carries significant implications across diverse domains, as it contributes to bolstering public safety within surveillance systems, ensuring safe navigation for autonomous robots and vehicles amidst human crowds, and enriching social awareness in human-robot interactions~\cite{wu2021comprehensive, chang2023importance, chang2022disentangling, martin2021jrdb, cl2021acoustic, chang2020transfer, ehsanpour2022jrdb, vendrow2023jrdb}. 
However, the advancement in related tasks is often impeded by the challenges of obtaining large-scale human group activity datasets in real-world scenarios with fine-grained multifaceted annotations.

Generating synthetic data is an emerging alternative to collecting real-world data due to its capability of producing large-scale datasets with perfect annotations. Nonetheless, most synthetic datasets \cite{varol2017learning, patel2021agora, ebadi2022psp, yang2023synbody, black2023bedlam} are primarily designed to facilitate human pose and shape estimation. 
They can only provide data with independently-animated persons, which is unsuitable for tasks in single-group and multi-group conditions~\cite{wu2021comprehensive}.
To address the limitation,  we propose {\datagenerator}, a synthetic data generator, with \textbf{m}ulti-view \textbf{m}ulti-group \textbf{m}ulti-person human \textbf{act}ions and group \textbf{act}ivities. 
As presented in Tab.~\ref{tab:dataset_comparison}, {\datagenerator} stands out by offering comprehensive annotations including both 2D and 3D annotations as well as fine-grained person-level and group-level labels, thereby making it an ideal synthetic dataset generator to support tasks such as human activity recognition and multi-person tracking across all listed real-world datasets. 

Illustrated in Fig.~\ref{fig:teaser}, our synthetic data generator features multiple semantic groups, highly diverse and photorealistic images, and a rich set of annotations. 
It encompasses 25 photometric 3D scenes, 104 HDRIs (High Dynamic Range Images), 5 lighting volumes, 2200 human models, 384 animations (categorized into 14 atomic action classes), and 6 group activities. 
For our experiments, We generated two datasets, \texttt{{\datagenerator}RGB} and \texttt{\texttt{{\datagenerator}3D}}. 
\texttt{{\datagenerator}RGB} contains both single-group and multi-group data with a total of 6M frames of RGB images and 48M bounding boxes, rendered in 20 FPS.  
\texttt{\texttt{{\datagenerator}3D}} is a 3D-only and single-group dataset, which contains 3D motions of all persons within a group. 
It has large group sizes (max 27 people) and an average of 6.7 persons per group. 
In total, \texttt{\texttt{{\datagenerator}3D}} contains a duration of 87.6 hours of group activities, captured in 30 FPS.

\begin{table*}[th]
\setlength{\tabcolsep}{6pt}
\footnotesize
  \aboverulesep=0ex
  \belowrulesep=0ex 
\begin{center}
\setlength{\tabcolsep}{5.6pt}
\begin{tabular}{lcccccc|cccc}
\toprule
\multirow{2}{*}{Dataset} & 
\multirow{2}{*}{\shortstack{Image\\Type}} & \multirow{2}{*}{\shortstack{Avatar\\Num.}} & \multirow{2}{*}{Video} & \multirow{2}{*}{\shortstack{Multi-\\View}} & \multirow{2}{*}{\shortstack{Multi-\\Person}}  &  \multirow{2}{*}{\shortstack{Multi-\\Group}}  & \multicolumn{4}{c}{Annotations}
\\
&&&&&&&2D&3D&Atomic Atn.&Group Act.
\\
\midrule

SURREAL, 2017 \cite{varol2017learning} & Composite & 145 & $\checkmark$ &  &  &  & $\checkmark$ & $\checkmark$ &  &  \\
AGORA, 2021 \cite{patel2021agora} & HDRI & 350 &  &  & $\checkmark$ &  & $\checkmark$ & $\checkmark$ &  &  \\
HSPACE, 2021 \cite{bazavan2021hspace} & 3D Scene & 1600 & $\checkmark$ & $\checkmark$ & $\checkmark$ &  & $\checkmark$ & $\checkmark$ &  &  \\
GTA-Humans, 2021 \cite{cai2021playing} & 3D Scene & 600 & $\checkmark$ &  & $\checkmark$ &  & $\checkmark$ & $\checkmark$ &  &  \\
PSP-HDRI+, 2022 \cite{ebadi2022psp} & HDRI & 28 &  &  & $\checkmark$ &  & $\checkmark$ &  & $\checkmark$ &  \\
SynBody, 2023 \cite{yang2023synbody} & 3D Scene & 10k & $\checkmark$ & $\checkmark$ &  $\checkmark$ &  & $\checkmark$ & $\checkmark$ &  &  \\
BEDLAM, 2023 \cite{black2023bedlam} & 3D Scene & 271 & $\checkmark$ &  & $\checkmark$ &  & $\checkmark$ & $\checkmark$ &  &  \\
\cellcolor{blue!15} \textbf{{\datagenerator} (Ours)} & \cellcolor{blue!15} Photometric 3D + HDRI & \cellcolor{blue!15} 2200 &  \cellcolor{blue!15} $\checkmark$ & \cellcolor{blue!15} $\checkmark$ & \cellcolor{blue!15} $\checkmark$ & \cellcolor{blue!15} $\checkmark$ & \cellcolor{blue!15} $\checkmark$ &  \cellcolor{blue!15} $\checkmark$ & \cellcolor{blue!15} $\checkmark$ & \cellcolor{blue!15} $\checkmark$ \\

\midrule

CAD, 2011 \cite{choi2011learning} & Real & - & $\checkmark$ &  & $\checkmark$ & $\checkmark$ & $\checkmark$ &  & $\checkmark$ & $\checkmark$ \\
Volleyball Dataset, 2016 \cite{ibrahim2016hierarchical} & Real & - & $\checkmark$ &  & $\checkmark$ & $\checkmark$ & $\checkmark$ &  & $\checkmark$ & $\checkmark$ \\
NTU-RGBD 120, 2019 \cite{liu2019ntu} & Real & - & $\checkmark$ & $\checkmark$ & $\checkmark$ &  & $\checkmark$ & $\checkmark$ & $\checkmark$ &  \\
\midrule

HiEve, 2020 \cite{lin2020human} & Real & - & $\checkmark$ &  & $\checkmark$ & $\checkmark$ & $\checkmark$ &  & $\checkmark$ &  \\
PoseTrack, 2021 \cite{doering2022posetrack21} & Real & - & $\checkmark$ &  & $\checkmark$ & $\checkmark$ & $\checkmark$ &  &  &  \\
MOT, 2020 \cite{dendorfer2020mot20} & Real & - & $\checkmark$ &  & $\checkmark$ & $\checkmark$ & $\checkmark$ &  &  &  \\
DanceTrack, 2022 \cite{sun2022dancetrack} & Real & - & $\checkmark$ &  & $\checkmark$ &  & $\checkmark$ &  &  &  \\
JRDB, 2023 \cite{martin2021jrdb, vendrow2023jrdb, ehsanpour2022jrdb} & Real & - & $\checkmark$ &  & $\checkmark$ & $\checkmark$ & $\checkmark$ &  & $\checkmark$ & $\checkmark$ \\

\bottomrule
\end{tabular}
\end{center}
\vspace{-15pt}
\caption{A comparison of synthetic datasets as well as commonly-used real datasets for activity understanding and person tracking. We refer to JRDB as a union set of JRDB, JRDB-Act, and JRDB-Pose datasets. Note that it offers 3D bounding boxes, but not poses.}
\vspace{-12pt}
\label{tab:dataset_comparison}
\end{table*}

We first demonstrate the merit of {\datagenerator} via synthetic data pre-training and mixed training on multi-person tracking and group activity recognition. 
For multi-person tracking, training with our synthetic data yields significant performance gain on several downstream methods \cite{zhang2023motrv2, zeng2022motr, MeMOTR, yan2023bridging}. 
We also demonstrate notable improvements in the state-of-the-art MOTRv2 method \cite{zhang2023motrv2} and observe that our synthetic data can substitute for 62.5\% more real-world data, without compromising performance.
In terms of group activity recognition, results indicate that pre-training with \texttt{{\datagenerator}RGB} greatly improves both person-level and group-level accuracy for Composer~\cite{zhou2022composer} and ActorTransformer~\cite{actortransformer} methods.
Based on our generated data, we then introduce a novel task, controllable 3D group activity generation, which aims to synthesize a group of 3D human motions, given control signals such as activity labels and group sizes. 
We systematically approach the new task by introducing both learning-based and heuristics-based metrics, along with a competitive baseline to generate meaningful human activities.


\noindent
This paper makes the following contributions:
\begin{itemize}
    \item 
    We propose a novel synthetic data generator, {\datagenerator}, and provide two large-scale synthetic datasets with highly diverse human activities, photorealistic multi-view videos, and comprehensive annotations.
    
    \item 
    We demonstrate that {\datagenerator} can significantly improve benchmark performances for multi-person tracking and group activity recognition and replace a large portion of real-world training data to reduce cost.
    
    \item 
    {\datagenerator} promotes new research initiatives for controllable 3D group activity generation, suggesting that synthetic data can not only support existing tasks but also create datasets for novel research.

\end{itemize}

\vspace{-8pt}
\section{Related Works}
\vspace{-5pt}
\noindent \textbf{Human Centered Synthetic Datasets.} 
The use of synthetic datasets for human-centered tasks has become increasingly prominent due to their diversity, scalability, and perfect annotations, with proven merits connected to various fields in machine learning, including domain adaptation~\cite{sun2022shift,liu2022learning}, heterogeneous multitask learning~\cite{yu2020bdd100k}, and sim2real~\cite{gao2022objectfolder2} or task2sim~\cite{mishra2022task2sim} transfer.
Most previous synthetic datasets are constructed to support human pose estimation. For example, SURREAL~\cite{varol2017learning} contains renderings of human motions from 145 avatars composited to a background image. Subsequent works \cite{patel2021agora, bazavan2021hspace, ebadi2022psp} managed to improve the image quality by leveraging realistic 3D scenes, high-quality renderings, and HDRI images.
Recently, synthetic datasets have been proposed to tackle human shape and mesh estimation. 
SynBody~\cite{yang2023synbody} constructs layered human assets to increase character diversity. 
BEDLAM~\cite{black2023bedlam} adds physically simulated hair and clothes to achieve state-of-the-art performances on shape and mesh estimation. 
Nonetheless, data with collective human motions and group activities cannot be obtained from them.
Our work, {\datagenerator}, is constructed with animated human groups tailored to \textit{multi-person} and \textit{multi-group} research.

\noindent \textbf{Real Multi-Person Datasets.}
Real-world datasets~\cite{choi2011learning, ibrahim2016hierarchical, liu2019ntu, lin2020human, dendorfer2020mot20, sun2022dancetrack, doering2022posetrack21, martin2021jrdb} with multiple persons are usually collected for tasks such as group activity understanding, multi-person tracking, and human trajectory prediction. 
Recognizing and parsing collective human activities \cite{zhou2022composer, actortransformer} rely primarily on multiple modalities (RGB, bounding box, pose) and hierarchical action and activity labels. 
These fine-grained labels are provided by datasets like CAD \cite{choi2011learning} and Volleyball Dataset \cite{ibrahim2016hierarchical}. 
On the other hand, datasets for person tracking, such as HiEve \cite{lin2020human}, MOT \cite{milan2016mot16, dendorfer2020mot20}, and DanceTrack \cite{sun2022dancetrack}, require not only 2D annotations (e.g., bounding box)  for individual frames, but also the association of the objects between them. 
Specifically, DanceTrack provides multiple persons in a group with the same clothing, making it difficult for the association of the individuals. 
MOT datasets target tracking for human crowds and contain mostly outdoor scenes from a bird-eye view. 
Recently, JRDB \cite{martin2021jrdb, ehsanpour2022jrdb, vendrow2023jrdb} with rich annotations is released. The images are captured by a social robot, navigating around daily scenes. 
It provides fine-grained annotations that support various tasks, including person detection, pose estimation, tracking, collective activity detection, and understanding.
{\datagenerator} not only offers the same modalities and annotations for supporting the aforementioned tasks, but it also provides full 3D annotations, making it suitable for a wide range of applications beyond the 2D domain. 

\vspace{-8pt}
\section{{\datagenerator}}
\vspace{-5pt}
{\datagenerator} is a \textbf{m}ulti-view \textbf{m}ulti-group \textbf{m}ulti-person human atomic \textbf{act}ion and group \textbf{act}ivity data generator built with Unity Engine and the Perception \cite{borkman2021unity} library. 
Inspired by PeopleSansPeople~\cite{ebadi2021peoplesanspeople} that populates randomly posed human avatars in a scene and renders static images, {\datagenerator} not only offers the same functionalities for human poses but also extends it to the spatio-temporal domain.
It generates RGB videos for dynamic human motions and produces a rich set of annotations simultaneously, including 
\textit{
(a) 2D and 3D joints/meshes, 
(b) 2D and 3D bounding boxes for individual persons, 
(c) atomic action and group activity categories,  
(d) tracking information such as individual and group IDs, 
(e) segmentation, depth, and normal images, and
(f) scene description.
}


\begin{figure*}[t]
\centering
   \includegraphics[width=\linewidth]{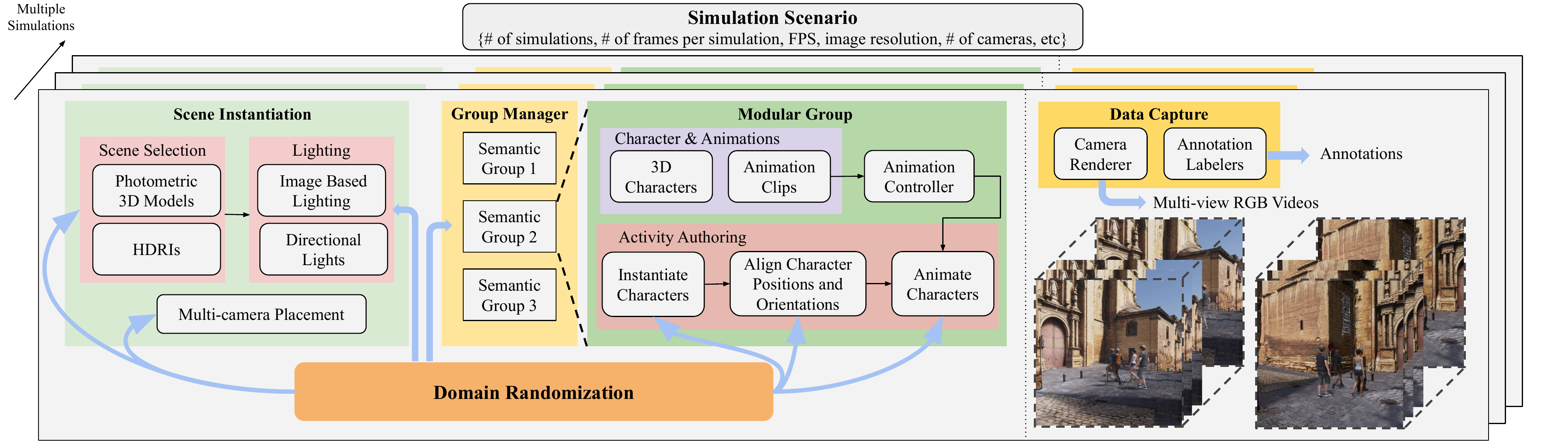}
   \vspace{-15pt}
   \caption{The data generation process of {\datagenerator}. It consists of multiple data simulations with scene instantiation, group activity authoring, and a data capture module. A high degree of randomization is involved in all aspects of the process to ensure diverse data. }
    \vspace{-12pt}
\label{fig:activity_pipeline}
\end{figure*}

\vspace{-3pt}
\subsection{Data Generation} 
\vspace{-5pt}

The process of our data generation is illustrated in Fig.~\ref{fig:activity_pipeline}. First, the generation process is configured by the simulation scenario that manages multiple independent simulations of human activities.
Then for each simulation, a 3D scene with background objects, lights, and cameras is set up, and groups of human characters are instantiated to be animated. 
Lastly, the multi-view RGB image frames are rendered and the annotations are exported at the end of the simulation.   


\noindent \textbf{Scene Instantiation.} We represent the environment through 25 photometric 3D scenes and 104 panoramic HDRIs. 
Each scene is initiated with randomized lighting and camera configuration. To attain a balance between realistic environmental illumination and pronounced shadow detail, our lighting schema integrates HDRI Sky lighting with a directional light. This directional light is subject to random variations in its direction, color temperature, and intensity. 
Regarding camera placement, we always point cameras towards the center of avatar groups and introduce variability by randomizing both the field of view and the camera’s distance to these groups. 


\noindent \textbf{Human Models and Motion Assets.}
{\datagenerator} leverages 2000 human models generated by Synthetic Humans \cite{picetti2023anthronet}, ranging across all ages (1 $\sim$100), genders,  ethnicities (such as Caucasian, Asian, Latin American, African, Middle Eastern), diverse body shapes, hair, and clothing. We also incorporated 200 widely-used human characters from 
RenderPeople \cite{renderpeople2023}. For the human motions, we collected 384 animation clips from AMASS~\cite{AMASS:ICCV:2019}, categorized into 14 atomic action classes. We created a universal animation controller that blends styles, including arm space and stride size, to create diverse motions from the collected clips.

\noindent \textbf{Modular Group Activities.} 
Each group activity is structured as a parameterized module, allowing for the customization of numerous variables. These variables include the number of individuals in the group and the specific atomic actions permitted within the group activity. 
This modularization ensures easy duplication, repositioning, and reuse of the group activities, enabling simulations of multiple groups at the same time. 
To procedurally animate a group of humans within a modular group, we establish the positions and orientations of the selected characters while choosing the appropriate animation clip for each character through the activity script. 
It's important to note that, despite drawing characters from the same set of avatars, the configurations and animations of these characters can vary significantly from one group to another. 
For example, animating a queueing activity may require all characters to be aligned in a straight line, while those in a walking group may form various shapes. The atomic actions that a person can perform also depend on the specific group activity. We carefully consider all these factors when authoring activities and provide a summary in the Sup. Mat.

\noindent \textbf{Domain Randomization.}
{\datagenerator} provides domain randomization for almost all aspects of the data generation process to ensure the simulation data is highly diverse. 
These aspects include 
the number of groups in a scene,
the number of persons in each group, 
the positions of groups, 
the alignment of persons in a group, 
the positions of individuals,
the textures for the instantiated characters,
and
the selection of scenes, lighting conditions, camera positions, characters, group activities, atomic actions, and animation clips.
Despite the fact that animating group activities inherently limits the degree of freedom in the placement of characters, by altering the shapes in which characters align (e.g, either in a cluster, a straight line, or a curve), {\datagenerator} nonetheless generates diverse activities and achieves sufficient randomization for downstream model generalization. 
More details regarding the randomization variables and their distributions are provided in Sup. Mat.

\noindent \textbf{Rendering and Annotations.}
{\datagenerator} utilizes the Unity high-definition render pipeline for the creation of photorealistic RGB images and leverages the Perception library for capturing annotations. 
On average, data is generated at a rate of 4.2 FPS using one NVIDIA RTX 3070 Ti graphics card, with a resolution of 1920x1080, and all annotations are enabled. Similar to PeopleSansPeople, the 2D skeleton follows COCO~\cite{lin2014microsoft} format, with additional labelers for exporting 3D joints, meshes, group IDs, and activity classes. After the data generation, the 3D joints and meshes are fitted with SMPL parameters~\cite{loper2015smpl, SMPL-X:2019}.

\vspace{-3pt}
\subsection{Dataset Statistics}
\vspace{-5pt}

{\datagenerator} comprises 25 photometric 3D scenes, 104 HDRIs, 5 lighting volumes, 2200 human models, 384 animations (categorized into 14 atomic action classes), and 6 group activities. 
Using the generator, we first generated our synthetic dataset, \texttt{{\datagenerator}RGB}. It contains 6K simulations of every single-group activity and 9K simulations of multi-group configuration, with 4 camera views. Each simulation produces a 5-second video clip, captured in 20 FPS and FHD resolution. 
In total, \texttt{{\datagenerator}RGB} contains 6M RGB images and 48M bounding boxes. The average track length is 4.65 seconds.

Additionally, we generated \texttt{{\datagenerator}3D}, a large-scale 3D-only dataset. It consists of more than 65K simulations of single-group activity. Each contains 150 frames of multi-person collective interactions in 30 FPS, resulting in a total duration of 87.6 hours. 
As shown in Tab.~\ref{tab:3d_datasets}, both the group size and the interaction complexity are significantly higher than those in previous multi-person motion datasets. 
Specifically, when compared with GTA-Combat \cite{song2022actformer}, {\datagenerator3D} contains more persons in a group, has more group activities, and provides a variable number of persons in a group. 
To the best of our knowledge, {\datagenerator3D} is the first large-scale 3D dataset for human group activities with large group sizes as well as per-frame individual action labels. 
See Sup. Mat. for detailed statistics of both datasets.

\begin{table}[!t] 
\setlength{\tabcolsep}{2.5pt}
\renewcommand{\arraystretch}{0.95}
\small
  \aboverulesep=0ex
  \belowrulesep=0ex 
\begin{center}
\begin{tabular}{l|c c | c c| c c}
\toprule
   \multirow{2}{*}{Dataset} 
   & \multirow{2}{*}{FPS}  & \# of & \multicolumn{2}{c|}{\# of Persons}  & \# of & \multirow{2}{*}{Duration} \\
     &  & Acty. & Avg & Max & Actn. &  \\
\midrule
   SBU \cite{yun2012two} &  15 & 8 & 2.0 & 2 & - & 7.6 mins \\
   Duet Dance \cite{kundu2020cross} & 25 & 5 & 2.0 & 2 & - & 2.3 hrs \\
   CHI3D \cite{fieraru2020three} & 50 & 8 & 2.0 & 2 & - & 2.7 hrs \\
   NTU RGBD 120 \cite{liu2019ntu} &  30 & \textbf{26} & 2.0 & 2 & - & 15.0 hrs \\
   GTA-Combat \cite{song2022actformer} &  15 & 1 & 3.2 & 5 & - & 39.0 hrs \\ \hline
   \texttt{{\datagenerator}3D}  &  30 & 6 & \textbf{6.7} & \textbf{27} & \textbf{8} & \textbf{87.6 hrs} \\
\bottomrule
\end{tabular}
\end{center}
\vspace{-15pt}
\caption{A comparison of datasets for 3D multi-person human activities. {\datagenerator3D} is the largest dataset with labels for atomic actions and more persons in a group.}
\vspace{-12.5pt}
\label{tab:3d_datasets}
\end{table}

\vspace{-5pt}
\section{Experiments}
\vspace{-5pt}
We showcase the practical utilities of {\datagenerator} through three core experiments: Multi-Person Tracking (MPT), Group Activity Recognition (GAR), and controllable Group Activity Generation (GAG).
The experiments are carefully designed to cover the following three perspectives:

\begin{itemize}
\setlength\itemsep{-0pt}
    \item[-] \textbf{Multi-modality}: Our experiments cover various modalities contained within our dataset, including RGB videos, 2D keypoints, and 3D joints. We leverage the rich annotations including bounding boxes, tracklets, group activities, and person action labels.
    
    
    \item[-] \textbf{Performance}: 
    We conduct the ablation study by altering the amount and the type of synthetic data used for training to see its effect on the model performance. 

    \item[-] \textbf{Novel task}: 
    We introduce a novel generative task (GAG), showing that synthetic data can not only support existing CV tasks but also create datasets for new research. 
\end{itemize}

\vspace{-2pt}
\subsection{Multi-Person Tracking} 
\label{sec:mpt_exp}
\vspace{-5pt}
The objective of MPT is to predict the trajectories of all persons from a dynamic video stream. 
Typically, person tracking involves two separate processes, person detection and association.
While the tracking task is approached in some prior works with the tracking-by-detection method \cite{bewley2016simple, wojke2017simple, cao2023observation}, we consider end-to-end approaches~\cite{zeng2022motr, zhang2023motrv2, yan2023bridging, MeMOTR} to evaluate the use of synthetic data on the performance of MPT as a whole, in lieu of an improved performance caused only by refined detection.

\noindent \textbf{Real-world Dataset: DanceTrack}~\cite{sun2022dancetrack} (DT) 
is a challenging MPT dataset characterized by dynamic movements with human subjects in uniform appearances. 
It has a total of 100 videos with over 105K frames.

\noindent \textbf{Synthetic Dataset}. 
Given the motion categories in the real-world dataset, we select a subset of \texttt{{\datagenerator}RGB} with groups of people dancing, walking, and running.
We use 1000 video clips with a single ``dance'' group as well as 1500 videos with a ``walk'' group and a ``run'' group simulated at the same time (denoted as WalkRun). 
We alter the use of the synthetic group activities (Dance, WalkRun, and Dance+WalkRun) in our experiments.

\begin{table}[t]
\setlength{\tabcolsep}{2.1pt}
\renewcommand{\arraystretch}{0.95}

\small
  \aboverulesep=0ex
  \belowrulesep=0ex 
    \begin{center}
    \begin{tabular}{l|c|c|c|c|c}
        \toprule
        Training Data & HOTA$\uparrow$ & DetA$\uparrow$ & AssA$\uparrow$ & IDF1$\uparrow$ & MOTA$\uparrow$ \\
        \midrule
        \footnotesize{DT$^{\circledast}$} & 69.8 & 83.0 & 58.9 & 71.6 & 89.3  \\
        \footnotesize{DT} & 68.8 \footnotesize{(10)} & 82.5 & 57.4 & 70.3 & 90.8  \\
        \footnotesize{DT + Syn (D)} & 59.0 & 75.5 & 46.1 & 59.0 & 82.6 \\
        \footnotesize{DT + Syn (WR)} & 70.1 & 83.1 & 59.4 & 72.5 & 92.0 \\
        \footnotesize{DT + Syn (WR+D)} & \textbf{71.9} \footnotesize{(2)} & \textbf{83.6} & \textbf{62.0} & \textbf{74.7} & \textbf{92.6} \\
        
        
        \footnotesize{DT + Syn$^{\dag}$ (WR+D)} & 72.2  & 83.4   & 62.6   & 75.5   & \textbf{92.7}   \\
        \footnotesize{DT (MOTRv2*)} & \textbf{73.4}  & \textbf{83.7}   & \textbf{64.4}   & \textbf{76.0}   & 92.1  \\

        \midrule
        

        
        \footnotesize{DT + BEDLAM~\cite{black2023bedlam}}   & 55.9 & 68.7 & 44.5 & 53.8 & 79.1 \\
        \footnotesize{DT + GTA-Humans~\cite{cai2021playing}} & 54.1 & 66.8 & 44.2 & 52.1 & 78.8 \\

        \bottomrule
    \end{tabular}
    \end{center}
    \vspace{-15pt}
    \caption{MPT results on DanceTrack with MOTRv2. ``D'' means synthetic dance group. ``WR'' means walk and run groups. ``WR+D'' refers to ``D'' and ``WR'' combined.
    The symbol $\circledast$ represents the author-provided checkpoint. 
    The symbol $\dag$ marks the same model with additional association at inference. Numbers in parentheses represent the rank in the DanceTrack leaderboard.}
    \vspace{-5pt}
    \label{tab:dt_result}   
\end{table}

\begin{table}[t]
\setlength{\tabcolsep}{2.5pt}
\renewcommand{\arraystretch}{0.95}

\small
  \aboverulesep=0.ex
  \belowrulesep=0.2ex 
    \begin{center}
    \begin{tabular}{lc|c|c|c|c|c}
        \toprule
         Model & Syn. Data & HOTA & DetA & AssA & IDF1 & MOTA \\

        \midrule
        
         &   & 54.2 & 73.5 & 40.2 & 51.5 & 79.7 \\
        \multirow{-2}{*}{\footnotesize MOTR~\cite{zeng2022motr}} & \checkmark  & \textbf{60.0} & \textbf{76.4} & \textbf{48.1} & \textbf{56.0} & \textbf{83.8} \\
        
        \midrule
        
         &  & 68.5 & 80.5 & 58.4 & 71.2 & 89.9 \\
        \multirow{-2}{*}{\footnotesize MeMOTR~\cite{MeMOTR}} & \checkmark  & \textbf{71.1} & \textbf{81.8} & \textbf{62.3} & \textbf{74.1} & \textbf{92.2} \\  
        
        \midrule
        
         &   & 69.4 & 82.1 & 58.9 & 71.9 & 91.2 \\
        \multirow{-2}{*}{\footnotesize CO-MOT~\cite{yan2023bridging}} & \checkmark  & \textbf{72.5} & \textbf{83.6} & \textbf{63.3} & \textbf{75.9} & \textbf{92.8} \\  
         
        \midrule

        &   &  68.8 & 82.5 & 57.4 & 70.3 & 90.8  \\
        \multirow{-2}{*}{\footnotesize MOTRv2~\cite{zhang2023motrv2}} & \checkmark  & \textbf{71.9} & \textbf{83.6} & \textbf{62.0} & \textbf{74.7} & \textbf{92.6} \\

        \midrule
        
                 &   & 73.4  & 83.7   & 64.4   & 76.0   & 92.1  \\
        \multirow{-2}{*}{\footnotesize MOTRv2*~\cite{zhang2023motrv2}} & \checkmark & \textbf{74.6} & \textbf{84.1}  & \textbf{64.9} & \textbf{76.4}  & \textbf{93.1} \\

        \bottomrule
    \end{tabular}
    \end{center}
    \vspace{-15pt}
    \caption{MPT results on DanceTrack using different methods trained with our synthetic data.}
    \vspace{-10pt}
    \label{tab:mpt_result_additional}   
\end{table}


\noindent \textbf{Results}. 
We mix together both synthetic and real data during training and present the results in Tab.~\ref{tab:dt_result}. 
First, adding our synthetic data yields significant improvement in all 5 tracking metrics as well as a hop in ranking on HOTA from 10$^{th}$ to 2$^{nd}$ place. 
The model trained with our synthetic data plus the extra association, marked as DT+Syn$^{\dag}$ (WR+D), achieves similar performance to MOTRv2*, the same model that is trained with additional validation data with an ensemble of 4 models~\cite{zhang2023motrv2}. 
This suggests that the synthetic data used in our experiment is equivalent to at least 62.5\% more real data.
Second, Compared with other synthetic data sources, such as BEDLAM and GTA-Humans, {\datagenerator} demonstrates superior performance, indicating its better suitability for multi-person dynamic conditions.
Third, we observe that the type of synthetic groups affects the model performance on real data. 
Adding the ``WalkRun'' groups to the training data is more effective than adding the ``Dance'' group. 
This is because while the DanceTrack dataset contains dynamic dance movements, the real challenge lies in detection and tracking when the subjects switch positions.
By design, the positions of the characters in our dance group are well-staged and the movements are nearly synchronous. (See Sup. Mat. for the design.)
Contrarily, having a walk and a run group together in a scene leads to frequent position switches relative to the camera view and thus improves the model performance. 
Lastly, Tab. \ref{tab:mpt_result_additional} presents the tracking results using different methods. Results indicate that our synthetic data is effective across various models.

\vspace{-2pt}
\subsection{Group Activity Recognition}
\label{sec:gar_exp}
\vspace{-5pt}
The goal of GAR is to determine the class of the group activity performed by the dominant group as well as the action class of each person. 
We consider \textbf{Composer~\cite{zhou2022composer}} and \textbf{Actor Transformer~\cite{actortransformer}} as the benchmark models. 
The former is a multi-scale transformer-based model and accepts only 2D keypoints as input. The latter can take combinations of multiple input modalities. 

\noindent \textbf{Real-world Dataset: CAD2 \cite{choi2011learning} and Volleyball Dataset \cite{ibrahim2016hierarchical}}.
CAD2 is an extended version of the Collective Activity Dataset~\cite{choi2009they} that records human group activities and is widely used for GAR benchmarks~\cite{wu2021comprehensive}. 
Volleyball Dataset (VD) is an action recognition dataset. It has 55 videos with 9 player action labels and 8 team activity labels. 

\noindent \textbf{Synthetic Dataset}. We use a subset of all single-group data from \texttt{{\datagenerator}RGB}. It contains a total of 10K videos and over 600K frames. 
It contains all group activity and individual action classes that CAD2 provides and further includes 7 more action types.

\begin{table}
\setlength{\tabcolsep}{2.7pt}
\setlength\extrarowheight{1.05pt}
\renewcommand{\arraystretch}{0.95}
\small
  \aboverulesep=0ex
  \belowrulesep=0ex 
\begin{center}
\begin{tabular}{c|c|c|c}
\toprule

\multirow{2}{*}{Model}  & \multirow{2}{*}{\shortstack{\footnotesize{Pretrained}\\\footnotesize{Syn. Data}}}  & \multicolumn{1}{c|}{Group Activity} & \multicolumn{1}{c}{Person Action} \\ 

\cline{3-4}
 &  &   Top 1 Acc (\%) $\uparrow$ & Top 1 Acc (\%) $\uparrow$ \\  
\midrule


\multirow{7}{*}{\shortstack{\textbf{Composer}\\\cite{zhou2022composer}}} &  N/A    &    $84.87^{\pm 2.3}$ (88.20)   &      $81.31^{\pm 2.4}$ (83.13) \\
\cline{2-4}
   & $10\%$ &      $86.12^{\pm 1.8}$ (87.87) &     $84.16^{\pm 1.8}$ (86.03)  \\
  &  $25\%$   &   $87.65^{\pm 1.2}$ (89.01) &    $86.36^{\pm 1.3}$ (86.81)  \\ 
   & $50\%$  &  $89.39^{\pm 0.4}$ (90.14) &   $86.68^{\pm 1.5}$ (87.99)  \\ 
  &  $100\%$   &    $\mathbf{89.74} ^{\pm 1.0}$ (\textbf{91.51}) &    $\mathbf{88.74} ^{\pm 1.7}$ (\textbf{89.05})   \\   
\cline{2-4}
 &    \textbf{Gains}  & \textbf{+4.87 (+3.31)} &  \textbf{+7.43 (+5.92)}  \\ 
\midrule 


\multirow{7}{*}{\shortstack{\textbf{Actor}\\ \textbf{Transformer}\\ \cite{actortransformer}}} &  N/A   &    $78.08^{\pm 1.0}$ (79.47)    &      $76.22^{\pm 2.2}$ (78.07)  \\
\cline{2-4}
  & $10\%$ &    $77.59^{\pm 2.4}$ (81.00)   &   $76.01^{\pm 3.2}$ (79.76)   \\
  &  $25\%$    &  $81.36^{\pm 2.1}$ (83.19)   &     $78.86^{\pm 2.4}$ (80.05)     \\   
  & $50\%$  &   $82.72^{\pm 1.3}$ (84.56)   &    $79.95^{\pm 1.6}$ (81.47)   \\ 
  &  $100\%$   &   $\mathbf{83.67}^{\pm 1.2}$ (\textbf{84.88})     &    $\mathbf{81.65}^{\pm 1.2}$ (\textbf{82.22})      \\   
\cline{2-4}
 &    \textbf{Gains}  & \textbf{+5.59 (+5.41)}  & \textbf{+5.43 (+4.15)} \\ 
 \bottomrule 
\end{tabular}
\end{center}
\vspace{-15pt}
\caption{Results of 2D keypoint-based \textbf{group activity and person action recognition} on CAD2 dataset. 
The best results are shown in parentheses.
The results suggest that pre-training with our synthetic data largely increases the accuracy for both group activity and person actions. 
Note that group accuracy saturates at 93.4\% and 86.2\% for Composer and Actor Transformer respectively.}
\label{tab:results_gar}
\vspace{-10pt}
\end{table}

\begin{table}[t]
\setlength{\tabcolsep}{2.0pt}
\renewcommand{\arraystretch}{0.95}

\small
  \aboverulesep=0ex
  \belowrulesep=0ex 
    \begin{center}
    \begin{tabular}{c|ccc|cc|cc}
        \toprule
        Model & 2D & RGB & Flow & CAD2 & Syn+CAD2 & VD & Syn+VD \\
        \midrule
        \textbf{Composer} & $\checkmark$ &  &  & 88.2 & 91.5  &  94.6 &  95.1  \\
        \midrule
        \multirow{5}{*}{\shortstack{\textbf{Actor}\\\textbf{Transformer}}} & $\checkmark$ &  &  & 79.5 & 84.9  & 92.3 & 93.7  \\
         &  & $\checkmark$ &  & 78.2 & 80.7  & 91.4 & 92.5  \\
         & $\checkmark$ & $\checkmark$ &  & 81.0 & \textbf{85.2}  & 93.5 & 94.3  \\        
         &  & $\checkmark$ & $\checkmark$ & 81.3 & 85.0  & 94.4 & \textbf{95.0}  \\
         & $\checkmark$ &  & $\checkmark$ & 79.5 & 81.9  & 93.0 & 94.1  \\
        \bottomrule
    \end{tabular}
    \end{center}
    \vspace{-15pt}
    \caption{The group activity recognition accuracy on CAD2  and Volleyball Dataset using different input modalities.}
    \label{tab:gar_result_additional}  
    \vspace{-10pt}

\end{table}


\noindent \textbf{Results}. 
We experimentally study how the size of our pre-training synthetic dataset and the capacity of a GAR model affect generalization from the synthetic to real domains.
We first train the models on different amounts of synthetic data.  
Then we fine-tune them on CAD2 and report the top 1 accuracy of both group activity and person action recognition on the test set of CAD2. 
Tab.~\ref{tab:results_gar} presents the results using only 2D keypoints as input. 
We see a common trend for both GAR models that the recognition accuracy increases as more synthetic data is used for pre-training.
With 100\% of synthetic data, the accuracy of Composer increases, with an average of 4.87\% at the group level and 7.43\% at the person level, whereas Actor Transformer sees a 5.59\% increase at group level and 5.43\% increase at person level.
Moreover, Tab. \ref{tab:gar_result_additional} shows the group recognition accuracy using different input modalities on CAD2 and VD. 
The performance gains in the experiment indicate that our synthetic data can effectively benefit the downstream GAR task across different methods, input modalities, and datasets.



\vspace{-5pt}
\subsection{Controllable 3D Group Activity Generation}
\label{sec:gag_exp}
\vspace{-5pt}
While the procedural generation of our group activities in {\datagenerator} yields realistic and diverse human activities, the implementation requires considerable effort and involves the design and application of specific heuristics.
Learning a generative model for human group activities, instead, encodes the heuristics to the architecture inherently and encompasses the capabilities of probabilistically generating diverse activities, with control over the entire group of human motions from various signals.
To this end, we introduce controllable 3D group activity generation (GAG).

\noindent \textbf{Definition.} Let $G_t^p = \{m_i^n\}_{i=1 \sim t, n = 1\sim p}$ be a group of human motions with $t$ frames and $p$ persons. 
The individual pose is denoted as $m_i^n \in R^{j\times d}$, where $j$ is 
the number of joints in the skeleton and $d$ is the joint dimension. 
The goal of GAG is to synthesize a group of 3D human motions $G_t^p$ from Gaussian noise, given an activity label and an arbitrary group size as input conditions. 
It requires a model capable of learning the temporal and spatial motion dependencies among persons within the same group and generating human motions with any group size simultaneously. 
GAG is related to dyadic motion generation \cite{chopin2023interaction, chang2022ivi} and partner-conditioned reaction generation \cite{rahman2022pacmo}, but involves the motion interactions of more than two persons. 



\noindent \textbf{Baselines.}
Although previous works \cite{chopin2023interaction, rahman2022pacmo, song2022actformer} can generate motions for multiple persons, they are limited to dyadic scenarios or groups with a fixed number of persons. 
Therefore, we present two baselines. 
%
\noindent The first one is the vanilla motion diffusion model, \textbf{MDM}~\cite{tevet2022human}. 
It was
proposed for probabilistic single-person motion generation from an input condition. We adopt their action-to-motion architecture for conditional synthesis and train the model on {\datagenerator3D} for generating an individual person's motion from an input group activity class label. 
In order to generate a group of human motions from a given group size, we repeat the single-person inference several times. 
In other words, the individual motions are generated independently by MDM. 
\noindent Our second baseline, \textbf{MDM+IFormer} is extended from MDM and includes an additional interaction transformer (IFormer) that works along the dimension of the persons. 
The interaction transformer encourages the model to learn the inter-person motion dependencies. 
At inference, MDM+IFormer is capable of producing coordinated group activities in one forward pass, due to its modeling of human interactions.

\noindent \textbf{Implementation.}
We utilize a common skeleton for all individual persons with 25 joints. We process the data so each motion is represented as both the 6D joint rotations \cite{zhou2019continuity} and the root positions. The final representation of a collective group activity with multiple persons is a tensor with shape (\#persons $\times$ \#frames $\times$ 26 $\times$ 6). For a fair comparison, both baseline models were trained on an NVIDIA RTX 3090 graphics card with 90\% data from \texttt{{\datagenerator}3D} for 320K iterations and then tested on the other 10\%.

\begin{table*}[!t]
\setlength{\tabcolsep}{3.6pt}
\small
  \aboverulesep=0ex
  \belowrulesep=0ex 
\begin{center}
\begin{tabular}{c|c c c c|c c c}
\toprule
   & \multicolumn{4}{c|}{Group Level} & \multicolumn{3}{c}{Person Level} \\
    \cline{2-8}

   & Acc $\uparrow$ & FID $\downarrow$ & Diversity $\rightarrow$ & Multimodality $\rightarrow$ & FID $\downarrow$ & Diversity $\rightarrow$ & Multimodality $\rightarrow$ \\ 
\midrule
   GT & $99.937$ & $0.001\pm0.000$ & $17.752\pm0.025$ & $3.491\pm0.012$ & $0.001\pm0.000$ & $14.506\pm0.013$ & $7.546\pm0.010$  \\
   MDM & $97.367$ & $3.909\pm0.019$ & $17.683\pm0.037$ & $\textbf{4.155}\pm0.019$ & $4.434\pm0.010$ & $14.158\pm0.035$ & $\textbf{7.588}\pm0.013$ \\
   MDM+IFormer & $\textbf{98.100}$ & $\textbf{3.242}\pm0.016$ & $17.855\pm0.040$ & $4.198\pm0.021$ & $\textbf{3.066}\pm0.007$ & $14.827\pm0.031$ & $6.945\pm0.011$ \\
\bottomrule
\end{tabular}
\end{center}
\vspace{-15pt}
\caption{The results of the generated group activities with the learning-based metrics at both levels. 
An up arrow means the result is better when the metric score is higher. A right arrow means the metric score should be close to ground truth (GT).} 

\label{tab:results_learningbased}
\vspace{-12pt}
\end{table*}

\vspace{-15pt}
\subsubsection{Evaluation}
\vspace{-5pt}

\noindent \textbf{Metrics.}
Due to the probabilistic nature of the task, we consider the following learning-based metrics, recognition accuracy, Frechet Inception Distance (FID), diversity, and multimodality, defined in \cite{guo2020action2motion}. 
These metrics, however, were originally designed for single-person motion generation. 
To evaluate the generated group activities, we report them at both group and person levels because they account for the fidelity and variations for the groups and the individuals. 
We train a multi-scale group activity recognition model using the Composer~\cite{zhou2022composer} architecture for the metrics.
See Sup. Mat. for detailed explanations of how to construct the learning-based metrics, including the recognition model as well as the latent representations at both levels.


\begin{figure}[t]
\centering
   \includegraphics[width=\linewidth]{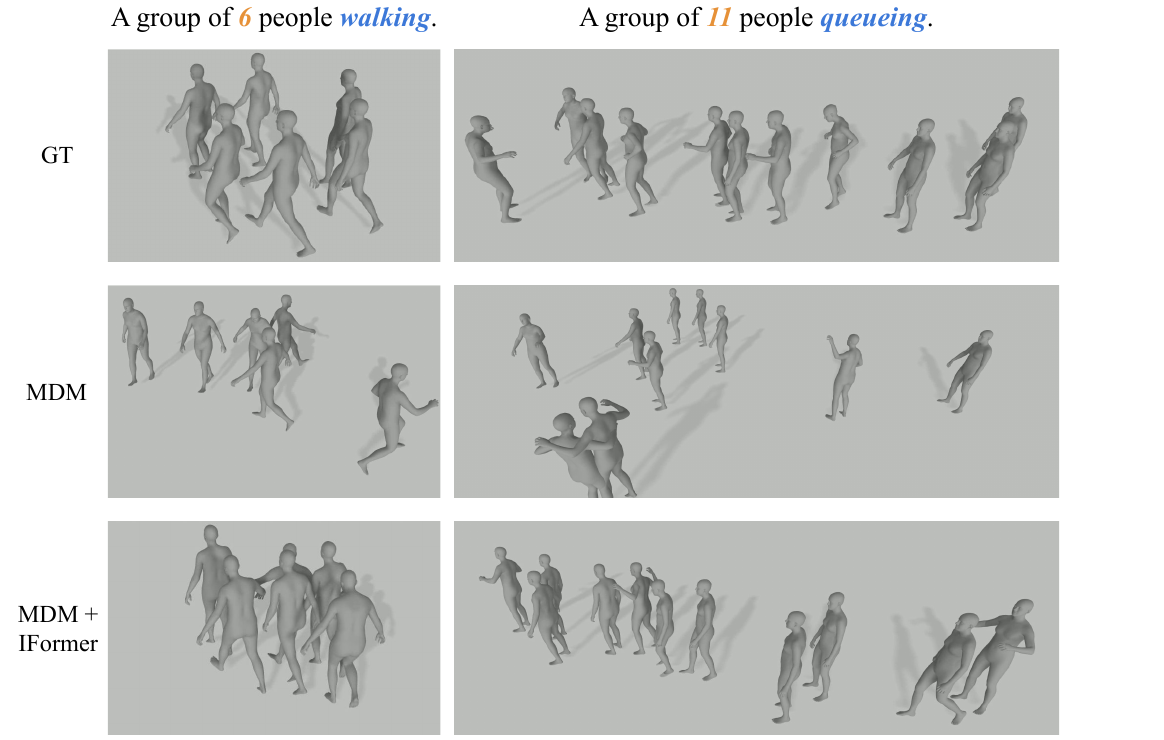}
   \vspace{-15pt}
   \caption{The qualitative comparison of two group activities from ground truth (GT), MDM, and MDM+IFormer. The distribution of the persons from MDM+IFormer is closer to GT.
   }
   \vspace{-10pt}
\label{fig:qualitative_3d_results}
\end{figure}

In addition to the learning-based metrics, we tailor four position-based metrics, \textit{collision frequency}, \textit{repulsive interaction force}, \textit{contact repulsive force}, and \textit{total repulsive force}, to the evaluation of human groups. 
The latter three are based on the social force model \cite{helbing1995social, helbing2000simulating}, which explains crowd behaviors using socio-psychological and physical forces. 
Here we describe the four metrics:

\noindent \textbf{\textit{Collision frequency}} indicates how often a collision (or an invalid interaction) would occur within a group. The collision count is calculated based on a distance threshold between any two persons in a group. It is then normalized by the total number of interactions to obtain the frequency. In other words, if $N$ persons are in a group, the normalization denominator is $N \cdot (N-1) / 2$.

\noindent \textbf{\textit{Repulsive interaction force}} describes the psychological tendency of two persons to stay away from each other. 
As the distance between two persons decreases, the repulsive force increases exponentially.

\noindent \textbf{\textit{Contact repulsive force}} represents the compression body force when two persons collide with each other. The contact force is nonzero only when two persons collide. A larger contact force means the interaction is less likely to occur.

\noindent \textbf{\textit{Total repulsive force}} is the accumulation of interaction and contact forces.

All four position-based metrics are calculated using the Euclidean distances of the persons' positions, with the shoulder width as the collision threshold. 
The social forces are calculated by averaging the magnitude of each individual's force accumulated through all its interactions with other persons. 
A well-performing model should generate group activities with low collision frequency and similar social force values to the ground truth.
For the evaluation, we generated 500 samples for each group activity using the two well-trained baseline models. Each generated sequence contains 60 frames. We use the test split as the ground truth and randomly extract the group activity of the same length for evaluation. 
To ensure the distributions of group sizes are similar to the ground truth, we calculate the minimum and maximum group sizes from the training split and uniformly sample a group size from that range to generate group activities.
Please refer to Sup. Mat. for more details regarding the baseline architectures, metrics formulas, and evaluation. 

\begin{table}[!t]
\setlength{\tabcolsep}{3.6pt}
\renewcommand{\arraystretch}{0.95}
\small
  \aboverulesep=0ex
  \belowrulesep=0ex 
\begin{center}
\begin{tabular}{c|cccc}
\toprule
    & Collision & Interaction & Contact & Total \\
    & Freq. $\downarrow$ & Force $\rightarrow$ & Force $\rightarrow$ & Force $\rightarrow$ \\
    \midrule
    GT & 0.037 & 65.79 & 46.55 & 112.33 \\
    MDM & 3.643 & 7,121.50 & 3,822.47 & 10,903.57 \\
    MDM+IFormer & \textbf{1.157} & \textbf{1,796.25} & \textbf{1,373.40} & \textbf{3167.80} \\
\bottomrule
\end{tabular}
\end{center}
\vspace{-15pt}
\caption{Results of the generated human activities with position-based metrics. The collision frequency is calculated on a 60-frame group activity and normalized by the total number of interactions in a group.} 
\vspace{-10pt}
\label{tab:results_positionbased}
\end{table}

\vspace{-12pt}
\subsubsection{Results}
\vspace{-5pt}

\noindent \textbf{MDM+IFormer is capable of generating group activities with well-aligned character positions.}
As shown in Fig.~\ref{fig:qualitative_3d_results}, 
MDM generates human groups that are poorly positioned. For example, the persons in a walking group do not walk in the same direction and the persons are poorly placed in a queueing group. 
This is because MDM generates the group activities by inferring the individual motions independently. 
The placement of the individuals simply follows the probabilistic distribution of all persons' positions in the dataset.
On the other hand, MDM+IFormer successfully learns the probabilistic distribution for the entire group due to its interaction transformer.
The persons are better aligned in a group and they have coordinated motions.

\noindent \textbf{Both baselines are capable of generating diverse activities that match the input condition, but MDM+IFormer obtains better FID scores.}
Tab.~\ref{tab:results_learningbased} shows the results for the learning-based metrics. 
When compared with ground truth, both baselines obtain similar scores on recognition accuracy, diversity, and multimodality at both levels. The results indicate that both models successfully learn to generate distinguishable individual motions and group activities. The generated motions are also as diverse as the ground truth.
The observations align with the results on action-to-motion generation in MDM \cite{tevet2022human}.
MDM+IFormer receives a lower FID score than MDM, suggesting that MDM+IFormer generates group activities with higher quality.


\noindent \textbf{The interaction transformer in MDM+IFormer greatly lowers the collision frequency within the generated group activities.} 
As shown in Tab.~\ref{tab:results_positionbased}, the collision frequency of the group activities generated by MDM+IFormer is much lower than the vanilla MDM. 
It suggests that the interaction transformer better learns the inter-person dependencies and generates more valid person interactions. 
In fact, we observe that the group activities generated by the vanilla MDM sometimes contain overlapping person positions. 
The high collision frequency of the MDM baseline also affects the repulsive forces, which makes social forces within the group activities of MDM implausible. 

\vspace{-8pt}
\section{Discussion and Conclusion} 
\vspace{-5pt}




We show the merit of {\datagenerator} by conducting three core experiments with multiple modalities and enhanced performances, as well as introducing a novel generative task.
In both MPT and GAR experiments, we observe positive correlations between the volume of synthetic data used for training and model performance, indicating an improved model generalizability to unseen test cases with more synthetic data. 
Moreover, our comparison between DT+Syn$^{\dag}$ and MOTRv2* reveals that synthetic data can replace certain real-world data from the target domain without sacrificing performance \cite{chang2024equivalency}.
Essentially, our synthetic data reduces the need for extensive real data during training, thereby effectively lowering the costs associated with data collection and annotation. 
This discovery represents a promising step towards achieving few-shot and potentially zero-shot sim2real transfer.
In our 3D Group Activity Generation experiment, we observe that MDM+IFormer, despite being a baseline for the novel task, learns to embed the heuristics for person interactions and produces well-aligned groups given the controls. 
It's important to highlight that the generative approach, though currently underperforms the procedural method (GT), demonstrates the unique potential of controlling the group motions directly from various signals, including activity class, group size, trajectory, density, speed, and text inputs. 
With the anticipation of more data availability and increased model capacity for generative models in the future, we expect the generative method to eventually prevail, leading to broader applications for social interactions and human collective activities.

While the complexity of group behaviors in our dataset may be constrained by the heuristics used for activity authoring, {\datagenerator} offers notable flexibility for incorporating new group dynamics tailored to any specific downstream tasks. These new groups could be derived from expert-guided heuristics, rules generated by large language models, or outputs from our 3D GAG model. 
Furthermore, we recognize the existing domain gaps between synthetic and real-world data. With more assets included in our data generator in future iterations, we can enhance model generalizability and alleviate the disparities.


\vspace{-8pt}
\section{Acknowledgement} 
\vspace{-5pt}
The research was supported in part by NSF awards: IIS-1703883, IIS-1955404, IIS-1955365, RETTL-2119265, and EAGER-2122119. This work was also partially supported by the Center for Smart Streetscapes, an NSF Engineering Research Center, under cooperative agreement EEC-2133516.
This material is based upon work supported by the U.S. Department of Homeland Security\footnote{Disclaimer. The views and conclusions contained in this document are those of the authors and should not be interpreted as necessarily representing the official policies, either expressed or implied, of the U.S. Department of Homeland Security.} under Grant Award Number 22STESE00001 01 01.

{
    \small
    \bibliographystyle{ieeenat_fullname}
    \bibliography{main}

\begin{thebibliography}{67}
\providecommand{\natexlab}[1]{#1}
\providecommand{\url}[1]{\texttt{#1}}
\expandafter\ifx\csname urlstyle\endcsname\relax
  \providecommand{\doi}[1]{doi: #1}\else
  \providecommand{\doi}{doi: \begingroup \urlstyle{rm}\Url}\fi

\bibitem[ren(2023)]{renderpeople2023}
Renderpeople.
\newblock \url{https://renderpeople.com/}, 2023.
\newblock Accessed: 2023-02-10.

\bibitem[Bazavan et~al.(2021)Bazavan, Zanfir, Zanfir, Freeman, Sukthankar, and Sminchisescu]{bazavan2021hspace}
Eduard~Gabriel Bazavan, Andrei Zanfir, Mihai Zanfir, William~T Freeman, Rahul Sukthankar, and Cristian Sminchisescu.
\newblock Hspace: Synthetic parametric humans animated in complex environments.
\newblock \emph{arXiv preprint arXiv:2112.12867}, 2021.

\bibitem[Bewley et~al.(2016)Bewley, Ge, Ott, Ramos, and Upcroft]{bewley2016simple}
Alex Bewley, Zongyuan Ge, Lionel Ott, Fabio Ramos, and Ben Upcroft.
\newblock Simple online and realtime tracking.
\newblock In \emph{2016 IEEE international conference on image processing (ICIP)}, pages 3464--3468. IEEE, 2016.

\bibitem[Black et~al.(2023)Black, Patel, Tesch, and Yang]{black2023bedlam}
Michael~J Black, Priyanka Patel, Joachim Tesch, and Jinlong Yang.
\newblock Bedlam: A synthetic dataset of bodies exhibiting detailed lifelike animated motion.
\newblock In \emph{Proceedings of the IEEE/CVF Conference on Computer Vision and Pattern Recognition}, pages 8726--8737, 2023.

\bibitem[Borkman et~al.(2021)Borkman, Crespi, Dhakad, Ganguly, Hogins, Jhang, Kamalzadeh, Li, Leal, Parisi, et~al.]{borkman2021unity}
Steve Borkman, Adam Crespi, Saurav Dhakad, Sujoy Ganguly, Jonathan Hogins, You-Cyuan Jhang, Mohsen Kamalzadeh, Bowen Li, Steven Leal, Pete Parisi, et~al.
\newblock Unity perception: Generate synthetic data for computer vision.
\newblock \emph{arXiv preprint arXiv:2107.04259}, 2021.

\bibitem[Cai et~al.(2021)Cai, Zhang, Ren, Wei, Ren, Lin, Zhao, Yang, Loy, and Liu]{cai2021playing}
Zhongang Cai, Mingyuan Zhang, Jiawei Ren, Chen Wei, Daxuan Ren, Zhengyu Lin, Haiyu Zhao, Lei Yang, Chen~Change Loy, and Ziwei Liu.
\newblock Playing for 3d human recovery.
\newblock \emph{arXiv preprint arXiv:2110.07588}, 2021.

\bibitem[Cao et~al.(2023)Cao, Pang, Weng, Khirodkar, and Kitani]{cao2023observation}
Jinkun Cao, Jiangmiao Pang, Xinshuo Weng, Rawal Khirodkar, and Kris Kitani.
\newblock Observation-centric sort: Rethinking sort for robust multi-object tracking.
\newblock In \emph{Proceedings of the IEEE/CVF Conference on Computer Vision and Pattern Recognition}, pages 9686--9696, 2023.

\bibitem[Chang(2020)]{chang2020transfer}
Che-Jui Chang.
\newblock Transfer learning from monolingual asr to transcription-free cross-lingual voice conversion.
\newblock \emph{arXiv preprint arXiv:2009.14668}, 2020.

\bibitem[Chang and Jeng(2021)]{cl2021acoustic}
Che-Jui Chang and Shyh-Kang Jeng.
\newblock Acoustic anomaly detection using multilayer neural networks and semantic pointers.
\newblock \emph{Journal of Information Science \& Engineering}, 37\penalty0 (1), 2021.

\bibitem[Chang et~al.(2022{\natexlab{a}})Chang, Zhang, and Kapadia]{chang2022ivi}
Che-Jui Chang, Sen Zhang, and Mubbasir Kapadia.
\newblock The ivi lab entry to the genea challenge 2022--a tacotron2 based method for co-speech gesture generation with locality-constraint attention mechanism.
\newblock In \emph{Proceedings of the 2022 International Conference on Multimodal Interaction}, pages 784--789, 2022{\natexlab{a}}.

\bibitem[Chang et~al.(2022{\natexlab{b}})Chang, Zhao, Zhang, and Kapadia]{chang2022disentangling}
Che-Jui Chang, Long Zhao, Sen Zhang, and Mubbasir Kapadia.
\newblock Disentangling audio content and emotion with adaptive instance normalization for expressive facial animation synthesis.
\newblock \emph{Computer Animation and Virtual Worlds}, 33\penalty0 (3-4):\penalty0 e2076, 2022{\natexlab{b}}.

\bibitem[Chang et~al.(2023)Chang, Sohn, Zhang, Jayashankar, Usman, and Kapadia]{chang2023importance}
Che-Jui Chang, Samuel~S Sohn, Sen Zhang, Rajath Jayashankar, Muhammad Usman, and Mubbasir Kapadia.
\newblock The importance of multimodal emotion conditioning and affect consistency for embodied conversational agents.
\newblock In \emph{Proceedings of the 28th International Conference on Intelligent User Interfaces}, pages 790--801, 2023.

\bibitem[Chang et~al.(2024)Chang, Li, Moon, and Kapadia]{chang2024equivalency}
Che-Jui Chang, Danrui Li, Seonghyeon Moon, and Mubbasir Kapadia.
\newblock On the equivalency, substitutability, and flexibility of synthetic data, 2024.

\bibitem[Choi et~al.(2009)Choi, Shahid, and Savarese]{choi2009they}
Wongun Choi, Khuram Shahid, and Silvio Savarese.
\newblock What are they doing?: Collective activity classification using spatio-temporal relationship among people.
\newblock In \emph{2009 IEEE 12th international conference on computer vision workshops, ICCV Workshops}, pages 1282--1289. IEEE, 2009.

\bibitem[Choi et~al.(2011)Choi, Shahid, and Savarese]{choi2011learning}
Wongun Choi, Khuram Shahid, and Silvio Savarese.
\newblock Learning context for collective activity recognition.
\newblock In \emph{CVPR 2011}, pages 3273--3280. IEEE, 2011.

\bibitem[Chopin et~al.(2023)Chopin, Tang, Otberdout, Daoudi, and Sebe]{chopin2023interaction}
Baptiste Chopin, Hao Tang, Naima Otberdout, Mohamed Daoudi, and Nicu Sebe.
\newblock Interaction transformer for human reaction generation.
\newblock \emph{IEEE Transactions on Multimedia}, 2023.

\bibitem[Dendorfer et~al.(2020)Dendorfer, Rezatofighi, Milan, Shi, Cremers, Reid, Roth, Schindler, and Leal-Taix{\'e}]{dendorfer2020mot20}
Patrick Dendorfer, Hamid Rezatofighi, Anton Milan, Javen Shi, Daniel Cremers, Ian Reid, Stefan Roth, Konrad Schindler, and Laura Leal-Taix{\'e}.
\newblock Mot20: A benchmark for multi object tracking in crowded scenes.
\newblock \emph{arXiv preprint arXiv:2003.09003}, 2020.

\bibitem[Doering et~al.(2022)Doering, Chen, Zhang, Schiele, and Gall]{doering2022posetrack21}
Andreas Doering, Di Chen, Shanshan Zhang, Bernt Schiele, and Juergen Gall.
\newblock Posetrack21: A dataset for person search, multi-object tracking and multi-person pose tracking.
\newblock In \emph{Proceedings of the IEEE/CVF Conference on Computer Vision and Pattern Recognition}, pages 20963--20972, 2022.

\bibitem[Ebadi et~al.(2021)Ebadi, Jhang, Zook, Dhakad, Crespi, Parisi, Borkman, Hogins, and Ganguly]{ebadi2021peoplesanspeople}
Salehe~Erfanian Ebadi, You-Cyuan Jhang, Alex Zook, Saurav Dhakad, Adam Crespi, Pete Parisi, Steven Borkman, Jonathan Hogins, and Sujoy Ganguly.
\newblock Peoplesanspeople: a synthetic data generator for human-centric computer vision.
\newblock \emph{arXiv preprint arXiv:2112.09290}, 2021.

\bibitem[Ebadi et~al.(2022)Ebadi, Dhakad, Vishwakarma, Wang, Jhang, Chociej, Crespi, Thaman, and Ganguly]{ebadi2022psp}
Salehe~Erfanian Ebadi, Saurav Dhakad, Sanjay Vishwakarma, Chunpu Wang, You-Cyuan Jhang, Maciek Chociej, Adam Crespi, Alex Thaman, and Sujoy Ganguly.
\newblock Psp-hdri $+ $: A synthetic dataset generator for pre-training of human-centric computer vision models.
\newblock \emph{arXiv preprint arXiv:2207.05025}, 2022.

\bibitem[Ehsanpour et~al.(2022)Ehsanpour, Saleh, Savarese, Reid, and Rezatofighi]{ehsanpour2022jrdb}
Mahsa Ehsanpour, Fatemeh Saleh, Silvio Savarese, Ian Reid, and Hamid Rezatofighi.
\newblock Jrdb-act: A large-scale dataset for spatio-temporal action, social group and activity detection.
\newblock In \emph{Proceedings of the IEEE/CVF Conference on Computer Vision and Pattern Recognition}, pages 20983--20992, 2022.

\bibitem[Fieraru et~al.(2020)Fieraru, Zanfir, Oneata, Popa, Olaru, and Sminchisescu]{fieraru2020three}
Mihai Fieraru, Mihai Zanfir, Elisabeta Oneata, Alin-Ionut Popa, Vlad Olaru, and Cristian Sminchisescu.
\newblock Three-dimensional reconstruction of human interactions.
\newblock In \emph{Proceedings of the IEEE/CVF Conference on Computer Vision and Pattern Recognition}, pages 7214--7223, 2020.

\bibitem[Gao and Wang(2023)]{MeMOTR}
Ruopeng Gao and Limin Wang.
\newblock {MeMOTR}: Long-term memory-augmented transformer for multi-object tracking.
\newblock In \emph{Proceedings of the IEEE/CVF International Conference on Computer Vision (ICCV)}, pages 9901--9910, 2023.

\bibitem[Gao et~al.(2022)Gao, Si, Chang, Clarke, Bohg, Fei-Fei, Yuan, and Wu]{gao2022objectfolder2}
Ruohan Gao, Zilin Si, Yen-Yu Chang, Samuel Clarke, Jeannette Bohg, Li Fei-Fei, Wenzhen Yuan, and Jiajun Wu.
\newblock Objectfolder 2.0: A multisensory object dataset for sim2real transfer.
\newblock In \emph{Proceedings of the IEEE/CVF Conference on Computer Vision and Pattern Recognition}, pages 10598--10608, 2022.

\bibitem[Gavrilyuk et~al.(2020)Gavrilyuk, Sanford, Javan, and Snoek]{actortransformer}
Kirill Gavrilyuk, Ryan Sanford, Mehrsan Javan, and Cees~GM Snoek.
\newblock Actor-transformers for group activity recognition.
\newblock In \emph{Proceedings of the IEEE/CVF Conference on Computer Vision and Pattern Recognition}, pages 839--848, 2020.

\bibitem[Ge et~al.(2021)Ge, Liu, Wang, Li, and Sun]{ge2021yolox}
Zheng Ge, Songtao Liu, Feng Wang, Zeming Li, and Jian Sun.
\newblock Yolox: Exceeding yolo series in 2021.
\newblock \emph{arXiv preprint arXiv:2107.08430}, 2021.

\bibitem[Guo et~al.(2020)Guo, Zuo, Wang, Zou, Sun, Deng, Gong, and Cheng]{guo2020action2motion}
Chuan Guo, Xinxin Zuo, Sen Wang, Shihao Zou, Qingyao Sun, Annan Deng, Minglun Gong, and Li Cheng.
\newblock Action2motion: Conditioned generation of 3d human motions.
\newblock In \emph{Proceedings of the 28th ACM International Conference on Multimedia}, pages 2021--2029, 2020.

\bibitem[Helbing and Molnar(1995)]{helbing1995social}
Dirk Helbing and Peter Molnar.
\newblock Social force model for pedestrian dynamics.
\newblock \emph{Physical review E}, 51\penalty0 (5):\penalty0 4282, 1995.

\bibitem[Helbing et~al.(2000)Helbing, Farkas, and Vicsek]{helbing2000simulating}
Dirk Helbing, Ill{\'e}s Farkas, and Tamas Vicsek.
\newblock Simulating dynamical features of escape panic.
\newblock \emph{Nature}, 407\penalty0 (6803):\penalty0 487--490, 2000.

\bibitem[Ho et~al.(2020)Ho, Jain, and Abbeel]{ho2020denoising}
Jonathan Ho, Ajay Jain, and Pieter Abbeel.
\newblock Denoising diffusion probabilistic models.
\newblock \emph{Advances in Neural Information Processing Systems}, 33:\penalty0 6840--6851, 2020.

\bibitem[Ibrahim et~al.(2016)Ibrahim, Muralidharan, Deng, Vahdat, and Mori]{ibrahim2016hierarchical}
Mostafa~S Ibrahim, Srikanth Muralidharan, Zhiwei Deng, Arash Vahdat, and Greg Mori.
\newblock A hierarchical deep temporal model for group activity recognition.
\newblock In \emph{Proceedings of the IEEE conference on computer vision and pattern recognition}, pages 1971--1980, 2016.

\bibitem[Kundu et~al.(2020)Kundu, Buckchash, Mandikal, Jamkhandi, Radhakrishnan, et~al.]{kundu2020cross}
Jogendra~Nath Kundu, Himanshu Buckchash, Priyanka Mandikal, Anirudh Jamkhandi, Venkatesh~Babu Radhakrishnan, et~al.
\newblock Cross-conditioned recurrent networks for long-term synthesis of inter-person human motion interactions.
\newblock In \emph{Proceedings of the IEEE/CVF winter conference on applications of computer vision}, pages 2724--2733, 2020.

\bibitem[Lin et~al.(2014)Lin, Maire, Belongie, Hays, Perona, Ramanan, Doll{\'a}r, and Zitnick]{lin2014microsoft}
Tsung-Yi Lin, Michael Maire, Serge Belongie, James Hays, Pietro Perona, Deva Ramanan, Piotr Doll{\'a}r, and C~Lawrence Zitnick.
\newblock Microsoft coco: Common objects in context.
\newblock In \emph{Computer Vision--ECCV 2014: 13th European Conference, Zurich, Switzerland, September 6-12, 2014, Proceedings, Part V 13}, pages 740--755. Springer, 2014.

\bibitem[Lin et~al.(2020)Lin, Liu, Liu, Li, Qian, Wang, Xu, Xiong, Qi, and Sebe]{lin2020human}
Weiyao Lin, Huabin Liu, Shizhan Liu, Yuxi Li, Rui Qian, Tao Wang, Ning Xu, Hongkai Xiong, Guo-Jun Qi, and Nicu Sebe.
\newblock Human in events: A large-scale benchmark for human-centric video analysis in complex events.
\newblock \emph{arXiv preprint arXiv:2005.04490}, 2020.

\bibitem[Liu et~al.(2019)Liu, Shahroudy, Perez, Wang, Duan, and Kot]{liu2019ntu}
Jun Liu, Amir Shahroudy, Mauricio Perez, Gang Wang, Ling-Yu Duan, and Alex~C Kot.
\newblock Ntu rgb+ d 120: A large-scale benchmark for 3d human activity understanding.
\newblock \emph{IEEE transactions on pattern analysis and machine intelligence}, 42\penalty0 (10):\penalty0 2684--2701, 2019.

\bibitem[Liu et~al.(2022)Liu, Kortylewski, Zhang, Li, Guo, Liu, Yuan, Mu, Qiu, and Yuille]{liu2022learning}
Qing Liu, Adam Kortylewski, Zhishuai Zhang, Zizhang Li, Mengqi Guo, Qihao Liu, Xiaoding Yuan, Jiteng Mu, Weichao Qiu, and Alan Yuille.
\newblock Learning part segmentation through unsupervised domain adaptation from synthetic vehicles.
\newblock In \emph{Proceedings of the IEEE/CVF Conference on Computer Vision and Pattern Recognition}, pages 19140--19151, 2022.

\bibitem[Loper et~al.(2015)Loper, Mahmood, Romero, Pons-Moll, and Black]{loper2015smpl}
Matthew Loper, Naureen Mahmood, Javier Romero, Gerard Pons-Moll, and Michael~J Black.
\newblock Smpl: A skinned multi-person linear model.
\newblock \emph{ACM transactions on graphics (TOG)}, 34\penalty0 (6):\penalty0 1--16, 2015.

\bibitem[Mahmood et~al.(2019)Mahmood, Ghorbani, Troje, Pons-Moll, and Black]{AMASS:ICCV:2019}
Naureen Mahmood, Nima Ghorbani, Nikolaus~F. Troje, Gerard Pons-Moll, and Michael~J. Black.
\newblock {AMASS}: Archive of motion capture as surface shapes.
\newblock In \emph{International Conference on Computer Vision}, pages 5442--5451, 2019.

\bibitem[Martin-Martin et~al.(2021)Martin-Martin, Patel, Rezatofighi, Shenoi, Gwak, Frankel, Sadeghian, and Savarese]{martin2021jrdb}
Roberto Martin-Martin, Mihir Patel, Hamid Rezatofighi, Abhijeet Shenoi, JunYoung Gwak, Eric Frankel, Amir Sadeghian, and Silvio Savarese.
\newblock Jrdb: A dataset and benchmark of egocentric robot visual perception of humans in built environments.
\newblock \emph{IEEE transactions on pattern analysis and machine intelligence}, 2021.

\bibitem[Milan et~al.(2016)Milan, Leal-Taix{\'e}, Reid, Roth, and Schindler]{milan2016mot16}
Anton Milan, Laura Leal-Taix{\'e}, Ian Reid, Stefan Roth, and Konrad Schindler.
\newblock Mot16: A benchmark for multi-object tracking.
\newblock \emph{arXiv preprint arXiv:1603.00831}, 2016.

\bibitem[Mishra et~al.(2022)Mishra, Panda, Phoo, Chen, Karlinsky, Saenko, Saligrama, and Feris]{mishra2022task2sim}
Samarth Mishra, Rameswar Panda, Cheng~Perng Phoo, Chun-Fu~Richard Chen, Leonid Karlinsky, Kate Saenko, Venkatesh Saligrama, and Rogerio~S Feris.
\newblock Task2sim: Towards effective pre-training and transfer from synthetic data.
\newblock In \emph{Proceedings of the IEEE/CVF Conference on Computer Vision and Pattern Recognition}, pages 9194--9204, 2022.

\bibitem[Patel et~al.(2021)Patel, Huang, Tesch, Hoffmann, Tripathi, and Black]{patel2021agora}
Priyanka Patel, Chun-Hao~P Huang, Joachim Tesch, David~T Hoffmann, Shashank Tripathi, and Michael~J Black.
\newblock Agora: Avatars in geography optimized for regression analysis.
\newblock In \emph{Proceedings of the IEEE/CVF Conference on Computer Vision and Pattern Recognition}, pages 13468--13478, 2021.

\bibitem[Pavlakos et~al.(2019)Pavlakos, Choutas, Ghorbani, Bolkart, Osman, Tzionas, and Black]{SMPL-X:2019}
Georgios Pavlakos, Vasileios Choutas, Nima Ghorbani, Timo Bolkart, Ahmed A.~A. Osman, Dimitrios Tzionas, and Michael~J. Black.
\newblock Expressive body capture: 3d hands, face, and body from a single image.
\newblock In \emph{Proceedings IEEE Conf. on Computer Vision and Pattern Recognition (CVPR)}, 2019.

\bibitem[Perez et~al.(2022)Perez, Liu, and Kot]{perez2022skeleton}
Mauricio Perez, Jun Liu, and Alex~C Kot.
\newblock Skeleton-based relational reasoning for group activity analysis.
\newblock \emph{Pattern Recognition}, 122:\penalty0 108360, 2022.

\bibitem[Picetti et~al.(2023)Picetti, Deshpande, Leban, Shahtalebi, Patel, Jing, Wang, au2, Sun, Laidlaw, Warren, Huynh, Page, Hogins, Crespi, Ganguly, and Ebadi]{picetti2023anthronet}
Francesco Picetti, Shrinath Deshpande, Jonathan Leban, Soroosh Shahtalebi, Jay Patel, Peifeng Jing, Chunpu Wang, Charles Metze~III au2, Cameron Sun, Cera Laidlaw, James Warren, Kathy Huynh, River Page, Jonathan Hogins, Adam Crespi, Sujoy Ganguly, and Salehe~Erfanian Ebadi.
\newblock Anthronet: Conditional generation of humans via anthropometrics.
\newblock 2023.

\bibitem[Rahman et~al.(2022)Rahman, Ghosh, Viswanath, Azizzadenesheli, and Bera]{rahman2022pacmo}
Md~Ashiqur Rahman, Jasorsi Ghosh, Hrishikesh Viswanath, Kamyar Azizzadenesheli, and Aniket Bera.
\newblock Pacmo: Partner dependent human motion generation in dyadic human activity using neural operators.
\newblock \emph{arXiv preprint arXiv:2211.16210}, 2022.

\bibitem[Song et~al.(2022)Song, Wang, Jiang, Fang, Ding, Gan, and Wu]{song2022actformer}
Ziyang Song, Dongliang Wang, Nan Jiang, Zhicheng Fang, Chenjing Ding, Weihao Gan, and Wei Wu.
\newblock Actformer: A gan transformer framework towards general action-conditioned 3d human motion generation.
\newblock \emph{arXiv preprint arXiv:2203.07706}, 2022.

\bibitem[Sun et~al.(2022{\natexlab{a}})Sun, Cao, Jiang, Yuan, Bai, Kitani, and Luo]{sun2022dancetrack}
Peize Sun, Jinkun Cao, Yi Jiang, Zehuan Yuan, Song Bai, Kris Kitani, and Ping Luo.
\newblock Dancetrack: Multi-object tracking in uniform appearance and diverse motion.
\newblock In \emph{Proceedings of the IEEE/CVF Conference on Computer Vision and Pattern Recognition}, pages 20993--21002, 2022{\natexlab{a}}.

\bibitem[Sun et~al.(2022{\natexlab{b}})Sun, Segu, Postels, Wang, Van~Gool, Schiele, Tombari, and Yu]{sun2022shift}
Tao Sun, Mattia Segu, Janis Postels, Yuxuan Wang, Luc Van~Gool, Bernt Schiele, Federico Tombari, and Fisher Yu.
\newblock Shift: a synthetic driving dataset for continuous multi-task domain adaptation.
\newblock In \emph{Proceedings of the IEEE/CVF Conference on Computer Vision and Pattern Recognition}, pages 21371--21382, 2022{\natexlab{b}}.

\bibitem[Tevet et~al.(2022)Tevet, Raab, Gordon, Shafir, Cohen-Or, and Bermano]{tevet2022human}
Guy Tevet, Sigal Raab, Brian Gordon, Yonatan Shafir, Daniel Cohen-Or, and Amit~H Bermano.
\newblock Human motion diffusion model.
\newblock \emph{arXiv preprint arXiv:2209.14916}, 2022.

\bibitem[Thilakarathne et~al.(2022)Thilakarathne, Nibali, He, and Morgan]{thilakarathne2022pose}
Haritha Thilakarathne, Aiden Nibali, Zhen He, and Stuart Morgan.
\newblock Pose is all you need: The pose only group activity recognition system (pogars).
\newblock \emph{Machine Vision and Applications}, 33\penalty0 (6):\penalty0 95, 2022.

\bibitem[Varol et~al.(2017)Varol, Romero, Martin, Mahmood, Black, Laptev, and Schmid]{varol2017learning}
Gul Varol, Javier Romero, Xavier Martin, Naureen Mahmood, Michael~J Black, Ivan Laptev, and Cordelia Schmid.
\newblock Learning from synthetic humans.
\newblock In \emph{Proceedings of the IEEE conference on computer vision and pattern recognition}, pages 109--117, 2017.

\bibitem[Vaswani et~al.(2017)Vaswani, Shazeer, Parmar, Uszkoreit, Jones, Gomez, Kaiser, and Polosukhin]{vaswani2017attention}
Ashish Vaswani, Noam Shazeer, Niki Parmar, Jakob Uszkoreit, Llion Jones, Aidan~N Gomez, {\L}ukasz Kaiser, and Illia Polosukhin.
\newblock Attention is all you need.
\newblock \emph{Advances in neural information processing systems}, 30, 2017.

\bibitem[Vendrow et~al.(2023)Vendrow, Le, Cai, and Rezatofighi]{vendrow2023jrdb}
Edward Vendrow, Duy~Tho Le, Jianfei Cai, and Hamid Rezatofighi.
\newblock Jrdb-pose: A large-scale dataset for multi-person pose estimation and tracking.
\newblock In \emph{Proceedings of the IEEE/CVF Conference on Computer Vision and Pattern Recognition}, pages 4811--4820, 2023.

\bibitem[Wojke and Bewley(2018)]{Wojke2018deep}
Nicolai Wojke and Alex Bewley.
\newblock Deep cosine metric learning for person re-identification.
\newblock In \emph{2018 IEEE Winter Conference on Applications of Computer Vision (WACV)}, pages 748--756. IEEE, 2018.

\bibitem[Wojke et~al.(2017)Wojke, Bewley, and Paulus]{wojke2017simple}
Nicolai Wojke, Alex Bewley, and Dietrich Paulus.
\newblock Simple online and realtime tracking with a deep association metric.
\newblock In \emph{2017 IEEE international conference on image processing (ICIP)}, pages 3645--3649. IEEE, 2017.

\bibitem[Wu et~al.(2021)Wu, Wang, Jian, Qiao, and Zhao]{wu2021comprehensive}
Li-Fang Wu, Qi Wang, Meng Jian, Yu Qiao, and Bo-Xuan Zhao.
\newblock A comprehensive review of group activity recognition in videos.
\newblock \emph{International Journal of Automation and Computing}, 18:\penalty0 334--350, 2021.

\bibitem[Yan et~al.(2023)Yan, Luo, Zhong, Gan, and Ma]{yan2023bridging}
Feng Yan, Weixin Luo, Yujie Zhong, Yiyang Gan, and Lin Ma.
\newblock Bridging the gap between end-to-end and non-end-to-end multi-object tracking, 2023.

\bibitem[Yang et~al.(2023)Yang, Cai, Mei, Liu, Chen, Xiao, Wei, Qing, Wei, Dai, et~al.]{yang2023synbody}
Zhitao Yang, Zhongang Cai, Haiyi Mei, Shuai Liu, Zhaoxi Chen, Weiye Xiao, Yukun Wei, Zhongfei Qing, Chen Wei, Bo Dai, et~al.
\newblock Synbody: Synthetic dataset with layered human models for 3d human perception and modeling.
\newblock \emph{arXiv preprint arXiv:2303.17368}, 2023.

\bibitem[Yu et~al.(2020)Yu, Chen, Wang, Xian, Chen, Liu, Madhavan, and Darrell]{yu2020bdd100k}
Fisher Yu, Haofeng Chen, Xin Wang, Wenqi Xian, Yingying Chen, Fangchen Liu, Vashisht Madhavan, and Trevor Darrell.
\newblock Bdd100k: A diverse driving dataset for heterogeneous multitask learning.
\newblock In \emph{Proceedings of the IEEE/CVF conference on computer vision and pattern recognition}, pages 2636--2645, 2020.

\bibitem[Yun et~al.(2012)Yun, Honorio, Chattopadhyay, Berg, and Samaras]{yun2012two}
Kiwon Yun, Jean Honorio, Debaleena Chattopadhyay, Tamara~L Berg, and Dimitris Samaras.
\newblock Two-person interaction detection using body-pose features and multiple instance learning.
\newblock In \emph{2012 IEEE computer society conference on computer vision and pattern recognition workshops}, pages 28--35. IEEE, 2012.

\bibitem[Zappardino et~al.(2021)Zappardino, Uricchio, Seidenari, and Del~Bimbo]{zappardino2021learning}
Fabio Zappardino, Tiberio Uricchio, Lorenzo Seidenari, and Alberto Del~Bimbo.
\newblock Learning group activities from skeletons without individual action labels.
\newblock In \emph{2020 25th International Conference on Pattern Recognition (ICPR)}, pages 10412--10417. IEEE, 2021.

\bibitem[Zeng et~al.(2022)Zeng, Dong, Zhang, Wang, Zhang, and Wei]{zeng2022motr}
Fangao Zeng, Bin Dong, Yuang Zhang, Tiancai Wang, Xiangyu Zhang, and Yichen Wei.
\newblock Motr: End-to-end multiple-object tracking with transformer.
\newblock In \emph{European Conference on Computer Vision}, pages 659--675. Springer, 2022.

\bibitem[Zhai et~al.(2022)Zhai, Hu, Yang, Zhou, and Liu]{zhai2022spatial}
Xiaolin Zhai, Zhengxi Hu, Dingye Yang, Lei Zhou, and Jingtai Liu.
\newblock Spatial temporal network for image and skeleton based group activity recognition.
\newblock In \emph{Proceedings of the Asian Conference on Computer Vision}, pages 20--38, 2022.

\bibitem[Zhang et~al.(2023)Zhang, Wang, and Zhang]{zhang2023motrv2}
Yuang Zhang, Tiancai Wang, and Xiangyu Zhang.
\newblock Motrv2: Bootstrapping end-to-end multi-object tracking by pretrained object detectors.
\newblock In \emph{Proceedings of the IEEE/CVF Conference on Computer Vision and Pattern Recognition}, pages 22056--22065, 2023.

\bibitem[Zhou et~al.(2022)Zhou, Kadav, Shamsian, Geng, Lai, Zhao, Liu, Kapadia, and Graf]{zhou2022composer}
Honglu Zhou, Asim Kadav, Aviv Shamsian, Shijie Geng, Farley Lai, Long Zhao, Ting Liu, Mubbasir Kapadia, and Hans~Peter Graf.
\newblock Composer: Compositional reasoning of group activity in videos with keypoint-only modality.
\newblock \emph{Proceedings of the 17th European Conference on Computer Vision (ECCV 2022)}, 2022.

\bibitem[Zhou et~al.(2019)Zhou, Barnes, Lu, Yang, and Li]{zhou2019continuity}
Yi Zhou, Connelly Barnes, Jingwan Lu, Jimei Yang, and Hao Li.
\newblock On the continuity of rotation representations in neural networks.
\newblock In \emph{Proceedings of the IEEE/CVF Conference on Computer Vision and Pattern Recognition}, pages 5745--5753, 2019.

\end{thebibliography}
}

\newpage
















\section*{A. Data Generator: {\datagenerator}}
\subsection*{A.1. Authoring of Group Activities}
Authoring group activities in {\datagenerator} is non-trivial because people adhere to social norms while forming groups.
The authoring requires nuanced adjustments varied from group to group, including the alignment of characters, their orientations, and the permitted atomic actions.
We summarize these rules and adjustments in Tab.~\ref{tab:authoring_groups}. 
For example, characters in a talking group are positioned in a circle facing the center. 
For queueing groups, characters can form a straight line, curve, etc., and individuals in a queue can be texting, idling, talking, and so on.

Group walking, running (jogging), and dancing are the three group activities with drastic body movements.
Particularly, collision avoidance is one social norm that is implicitly followed by humans during collective walking or running. Therefore,  we propose a simple algorithm to dynamically adjust the animation speed for each character engaged in the walking and running activities, thus mitigating avatar collisions. 
The detailed algorithm is shown in Alg.~\ref{alg:collision_avoidance}. 
Specifically, the animation speed of a character decreases when any character is in front of it and close to it. 
When no potential collision is detected, the animation speed of a character can increase up to its initial speed.
While animating complex interactions for group dancing is challenging, we enforce nearly synchronous movements for all individuals.
\begin{table*}[]
\setlength{\tabcolsep}{8.6pt}
\footnotesize
  \aboverulesep=0ex
  \belowrulesep=0ex 
\begin{center}
\setlength{\tabcolsep}{5.6pt}
\begin{tabular}{c|c|c|c|c}
\toprule
\textbf{Activity Name} & \textbf{Alignment} & \textbf{Face At} & \textbf{Atomic Actions} & \textbf{Other Conditions} \\
\midrule \midrule
\multirow{2}{*}{Walking} & \multirow{2}{*}{Straight Line, Circle, Rectangle} & \multirow{2}{*}{Same Direction} & \multirow{2}{*}{walk} &  \multirow{2}{*}{Adjust animation speed at runtime} \\
 & & & & \\
\hline
\multirow{2}{*}{Waiting} & \multirow{2}{*}{Multi-Row Straight Lines} & \multirow{2}{*}{Same Direction} & \multirow{2}{*}{idle, text, talk, point, wave} & \multirow{2}{*}{N/A} \\
 & & & & \\

\hline

\multirow{2}{*}{Queueing} & \multirow{1}{*}{Straight Line, One-Corner Line,} & \multirow{2}{*}{Front of Queue} & \multirow{2}{*}{idle, text, talk, point} & \multirow{2}{*}{N/A} \\
 & Two-Corner Line, Parabola, Curve & & & \\
\hline

\multirow{2}{*}{Talking} & \multirow{2}{*}{Circle} & \multirow{2}{*}{Group Center} & \multirow{2}{*}{talk} & \multirow{2}{*}{N/A} \\
 & & & & \\
\hline

\multirow{2}{*}{Dancing} & \multirow{2}{*}{Multi-Row Straight Lines} & \multirow{2}{*}{Same Direction} & \multirow{2}{*}{dance} & \multirow{2}{*}{Nearly-synchronous movements}  \\
& & & & \\

\hline

\multirow{2}{*}{Jogging} & \multirow{2}{*}{Straight Line, Circle, Rectangle} & \multirow{2}{*}{Same Direction} & \multirow{2}{*}{run} & \multirow{2}{*}{Adjust animation speed at runtime} \\
 & & & & \\
\bottomrule
\end{tabular}
\end{center}
\vspace{-10pt}
\caption{Rules and adjustments for activity authoring.}
\label{tab:authoring_groups}
\end{table*}

\begin{table*}[]
\setlength{\tabcolsep}{6.6pt}
\footnotesize
  \aboverulesep=0ex
  \belowrulesep=0ex 
\begin{center}
\setlength{\tabcolsep}{5.6pt}
\begin{tabular}{c|c|c|c}
\toprule
\textbf{Category} & \textbf{Randomizer} & \textbf{Variable} & \textbf{Distribution} \\
\midrule
\midrule
\multirow{2}{*}{Scenes} & Scene Selection  & scene & a set of prebuilt 3D environments \\
\cline{2-4}
 & HDRI Selection & hdri & a set of collected HDRIs
 \\

\midrule
\multirow{4}{*}{Camera} & \multirow{4}{*}{Camera Position } & radius & Uniform(6, 10) \\
 &  & camera rotation & Uniform(0, 360) \\
 &  & camera height & Uniform(1, 5) \\
 &  & perturbation & Cartesian[Uniform(-1, 1), 0 Uniform(-1, 1)] \\

\midrule
\multirow{6}{*}{Lights} & Lighting Volume & volume & a set of lighting volume conditions \\
\cline{2-4}
 & Light Type  & light type & a set of light types \\
\cline{2-4}
 & \multirow{2}{*}{Light Position } & XZ position & Cartesian[Uniform(-20, 20), 0, Uniform(-20, 20)] \\
 &  & height & Uniform(5, 10) \\
\cline{2-4}
 & Light Intensity  & intensity range & Uniform(0.5, 3) \\
\cline{2-4}
 & Light Rotation  & orientation range & Face at Cartesian[Uniform(-50, 50), 0 Uniform(-50, 50)] \\

\midrule
\multirow{4}{*}{Multi-Group} & Group Number  & group number range & UniformRange(1, MaxNumGroups) \\
\cline{2-4}
 & Group Selection  & group activity & a set of modular groups \\
\cline{2-4}
 & \multirow{2}{*}{Group Placement } & group position & Cartesian[Uniform(-20, 20), 0, Uniform(-20, 20)]
 \\
 &  & group rotation & Euler[0, Uniform(0, 360), 0] \\
 
\midrule
\multirow{14}{*}{\shortstack{Activity\\Authoring}} & Character Number  & character number range & UniformRange(1, MaxNumCharacters) \\
\cline{2-4}
 & Multi-person Subgroup  & multi-person number range & UniformRange(0, MaxNum) \\
\cline{2-4}
 & Character Selection  & character & a set of 2200 characters \\
\cline{2-4}
 & \multirow{2}{*}{Character Texture } & body color & RGBA[Uniform(0.4, 1), Uniform(0.4, 1), Uniform(0.4, 1), Uniform(0.6, 1)] \\
 &  & clothes colors & HSV[Uniform(0, 1), Uniform(0, 1), Uniform(0.4, 1)] \\
\cline{2-4}
 & Character Alignment  & alignment method & a set of aligment methods \\
\cline{2-4}
 & Character Interval  & interval & Uniform(MinInterval, MaxInterval) \\
\cline{2-4}
 & \multirow{2}{*}{Character Perturbation } & position perturbation & Uniform(-0.25*interval, 0.25*interval)  \\
 &  & rotation perturbation & Uniform(-45$^{\circ}$, 45$^{\circ}$) \\
\cline{2-4}
 & Atomic Action  & atomic action & a set of permitted atomic actions \\
\cline{2-4}
 & \multirow{4}{*}{Animation } & animation clip & a set of animation clips \\
 &  & blended parameter & Uniform(0, 1) \\
 &  & speed & Uniform(0.8, 1.2) \\
 &  & normalized starting time & Uniform(0, 1) \\

\bottomrule
\end{tabular}
\end{center}
\vspace{-10pt}
\caption{List of randomizers, variables, and distributions used in  {\datagenerator}.}
 \vspace{-10pt}
\label{tab:domain_randomization}
\end{table*}

\subsection*{A.2. List of Variables for Domain Randomization}
\textit{Domain Randomization} allows {\datagenerator} to generate massive-scale diverse group activity data. 
Compared to PeopleSansPeople~\cite{ebadi2021peoplesanspeople},~{\datagenerator} contains a much higher degree of domain randomization for animating human motions and activities.
{\datagenerator} consists of a total of 14 atomic action classes and 384 animation clips, each with several blended style parameters such as character arm-space and stride. 
The domain randomization covers the scenes, cameras, lights, multi-groups, and activity authoring, as listed in Tab.~\ref{tab:domain_randomization}. We describe the randomizers in {\datagenerator} below.

\begin{algorithm}[t]
\caption{Dynamic speed adjustment for collision avoidance.}\label{alg:collision_avoidance}
\begin{algorithmic}[1]
\Require a list of instantiated characters and their initial animation speeds. 
\State At every frame,
\For{\textit{character} in Characters}
    \State \textit{init\_speed} $\gets$ Initial Speed of \textit{character}
    \State \textit{Speed} $\gets$ Current Speed of \textit{character}
    \State \textit{pos} $\gets$ Position of \textit{character} 
    \State \textit{forward} $\gets$ Forward Vector of \textit{character} 

    \State \textit{flag} $\gets$ False
    \For{\textit{other\_character} in Characters}
        \If{\textit{character} is not \textit{other\_character}}
            \State \textit{pos\_other} $\gets$ Position of \textit{other\_character} 
            \State \textit{offset} $\gets$ \textit{pos\_other} - \textit{pos}
            \State \textit{angle} $\gets$ Angle Between \textit{forward} and \textit{offset}
            \State \textit{dist} $\gets$ Length of \textit{offset}

            \If{\textit{dist} $\leq$ 0.8 $\And$ \textit{angle} $\leq$ 60}
                \State \textit{flag} $\gets$ True
            \EndIf
        \EndIf
    \EndFor
    \If{\textit{flag}}
            \State\textit{Speed} $\gets$ Max(\textit{Speed} * 0.96, 0.1) 
        \Else
            \State \textit{Speed} $\gets$ Min(\textit{Speed} * 1.03, \textit{init\_speed}) 

    \EndIf
\EndFor

\end{algorithmic}
\end{algorithm}

\begin{itemize}
\setlength\itemsep{-2pt}
\item \textbf{Scene Selection Randomizer} randomizes the selection of 3D scene.

\item  \textbf{HDRI Randomizer} randomizes the selection of panorama HDRIs.

\item  \textbf{Camera Position Randomizer} includes the randomizations of camera height, distance, and angle in a cylindrical coordinate.

\item  \textbf{Light Type Randomizer} randomizes the light type.

\item  \textbf{Light Position Randomizer} randomizes the positions of all lights. 

\item  \textbf{Light Intensity Randomizer} randomizes the intensities of all lights.

\item  \textbf{Light Rotation Randomizer} randomizes the rotations of all lights.

\item  \textbf{Group Number Randomizer} randomizes the number of groups being instantiated during the simulation.

\item  \textbf{Group Selection Randomizer} randomizes the selection of the activity for each group in the scene.

\item  \textbf{Group Placement Randomizer} randomizes the center position for each group.

\item  \textbf{Character Number Randomizer} randomizes the number of characters being instantiated in a group.

\item  \textbf{Multi-person Subgroup Randomizer} randomizes the number of 
\textit{subgroups} in an activity, such as two persons talking to each other in a queueing group. This randomizer 
applies to queueing and waiting groups.

\item  \textbf{Character Selection Randomizer} randomizes the selection of characters.

\item  \textbf{Character Texture Randomizer} randomizes the clothes and body colors of all characters.

\item  \textbf{Character Alignment Randomizer}  randomizes the method used to align characters in a group.

\item \textbf{Character Interval Randomizer} randomizes the interval between characters.

\item \textbf{Character Perturbation Randomizer} adds small perturbations to the characters' positions and rotations.

\item  \textbf{Atomic Action Randomizer}  randomizes the selection of permitted atomic actions.

\item  \textbf{Animation Randomizer} includes the randomization of animation clips, blended style parameters, animation speeds, and playback offsets.
\end{itemize}

\section*{B. Datasets}
\begin{figure}[t]
\centering
   \includegraphics[width=\linewidth]{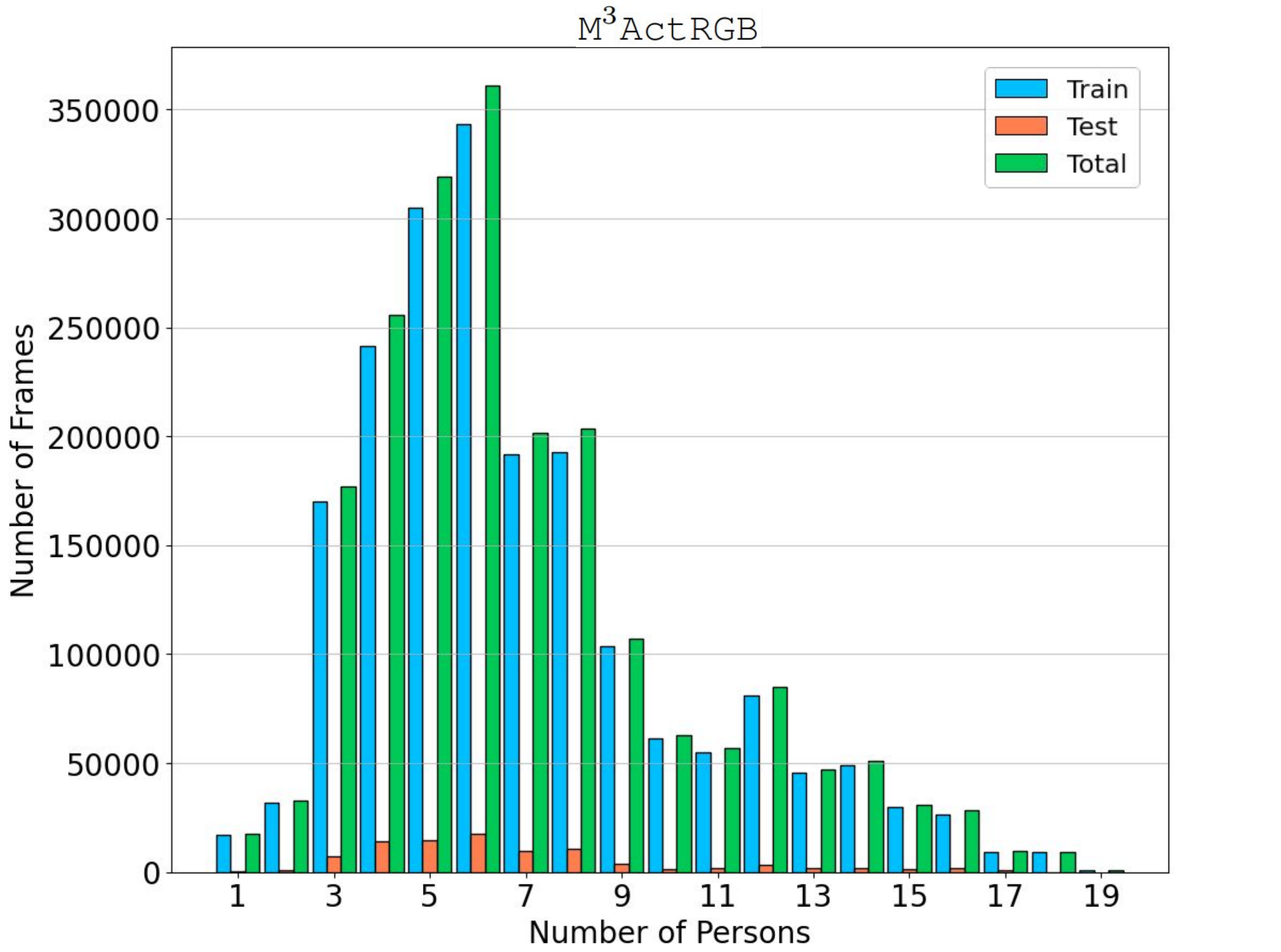}
   \caption{Distributions of \texttt{{\datagenerator}RGB}.}
\label{fig:m3actrgb_distr}
\end{figure}

\begin{figure*}[t]
\centering
   \includegraphics[width=\linewidth]{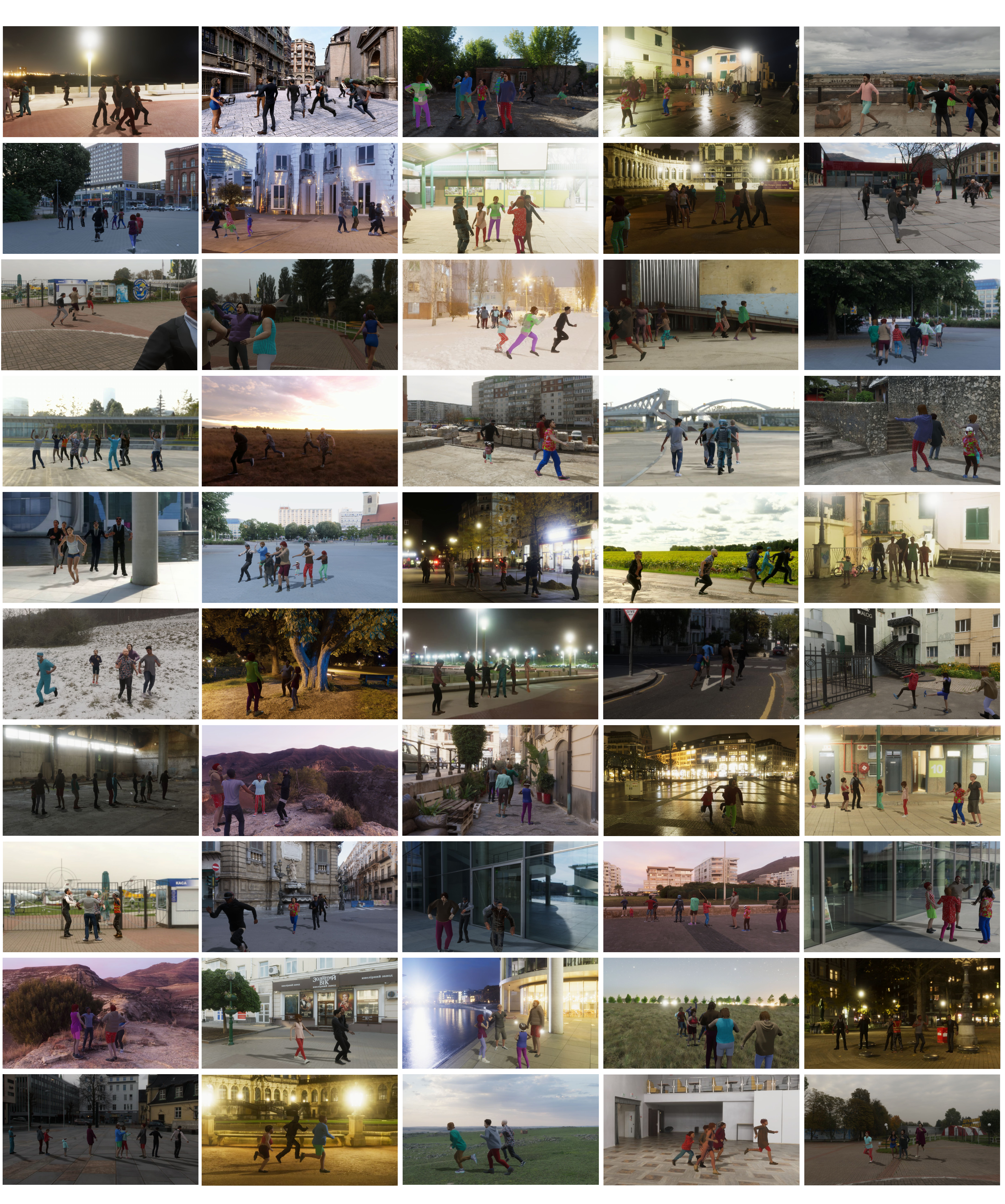}
   \caption{Collage of images from \texttt{{\datagenerator}RGB}, including multi-group activities (first 3 rows) and single-group activities (last 7 rows).}
\label{fig:m3actrgb_images}
\end{figure*}

\noindent \textbf{\texttt{{\datagenerator}RGB}} contains 9K videos of multi-group and 6K videos of single-group activities, with a total of 6M RGB images and 48M bounding boxes. 
We show a collage of images from \texttt{{\datagenerator}RGB} in Fig.~\ref{fig:m3actrgb_images}, which contains diverse and realistic multi-group and single-group activities. 
(See our supplementary video for animated data samples.)
The distribution of \texttt{{\datagenerator}RGB} is also shown in Fig.~\ref{fig:m3actrgb_distr}. \texttt{{\datagenerator}RGB} contains as many as 19 persons per frame and an average of 8.1 persons per frame. 
Additionally, we show the comparison of several tracking datasets in Tab.~\ref{tab:tracking_dataset_comparison}. 
Even though we only selected the ``WalkRun'' and ``Dance'' data from \texttt{{\datagenerator}RGB} for the tracking experiments, the dataset size is much larger compared to MOT17~\cite{milan2016mot16} and DanceTrack~\cite{sun2022dancetrack}. 
In terms of the average number of tracks per video, our dataset is closer to DanceTrack. MOT17 mostly contains crowded scenes, while DanceTrack has only one dancing group per video.

\begin{table}[!t] 
\setlength{\tabcolsep}{2.5pt}
\small
  \aboverulesep=0ex
  \belowrulesep=0ex 
\begin{center}
\begin{tabular}{l|l|l|l}
\toprule
 Dataset & MOT17 \cite{milan2016mot16} & DanceTrack \cite{sun2022dancetrack} & \texttt{{\datagenerator}RGB}$^{\dag}$ \\
\midrule
 \#Videos & 14 & 100 & \textbf{2500} \\
 Avg. \#Tracks & \textbf{56} & 9 & 8 \\
 Avg. Track Len. (s) & 35.4 & \textbf{52.9} & 4.7 \\
 FPS (s) & 30 & 20 & 20 \\
 Total Frames & 11,235 & 105,855 & \textbf{250,000} \\
\bottomrule
\end{tabular}
\end{center}
\vspace{-10pt}
\caption{Comparison of multi-object tracking datasets. \texttt{{\datagenerator}RGB}$^{\dag}$ consists of ``WalkRun'' and ``Dance'' data used in our tracking experiments.}
\label{tab:tracking_dataset_comparison}
\end{table}

\begin{figure}[t]
\centering
   \includegraphics[width=\linewidth]{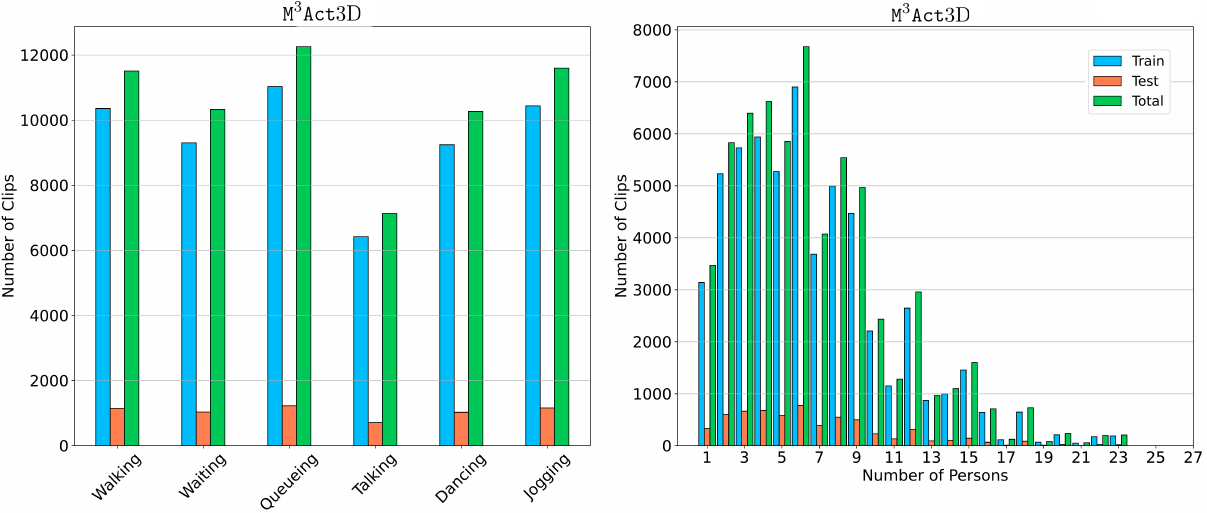}
   \caption{Distributions of \texttt{{\datagenerator}3D}.}
\label{fig:m3act3d_distr}
\end{figure}

\noindent \textbf{\texttt{{\datagenerator}3D}} has 65K simulations of 3D single-group motions with a total duration of 87.6 hours, captured in 30 FPS.
Unlike \texttt{{\datagenerator}RGB} which contains equally simulated group activities, \texttt{{\datagenerator}3D} has different data sizes of all semantic groups based on their complexity.
The complexity of a group includes its alignment methods, permitted atomic actions, animation clips, and styles.
Fig.~\ref{fig:m3act3d_distr} shows the distribution of \texttt{{\datagenerator}3D}. 
Specifically, \texttt{{\datagenerator}3D} has more queueing groups than talking groups because the persons can form various shapes and more atomic actions can be performed within a queueing group.
We also slightly increased the range of the number of persons in the group for \texttt{{\datagenerator}3D}. 
On average, it has 6.7 persons for every single group and a maximum of 27 persons.

\section*{C. More Details of Experiments}
\subsection*{C.1 Multi-Person Tracking} 
The goal of multi-person tracking (MPT) is to predict the trajectories (bounding boxes + identification) of all persons across an image sequence from a dynamic video stream. 
Traditionally, multi-object tracking is approached by adding a re-identification layer, either using trainable architecture~\cite{wojke2017simple, Wojke2018deep} or applying heuristics-based algorithms~\cite{bewley2016simple}, on top of the object detection results, aiming to associate the bounding boxes across frames.
Recently, end-to-end methods~\cite{zeng2022motr, zhang2023motrv2} have shown to be more effective in several challenging datasets, such as DanceTrack~\cite{sun2022dancetrack} and MOT17~\cite{milan2016mot16}. 
To demonstrate the effectiveness of our synthetic data in enhancing real-world performance in multi-person tracking, we assess the impact primarily on MOTRv2~\cite{zhang2023motrv2}. 
It is an extension of MOTR~\cite{zeng2022motr} by incorporating YOLO-X~\cite{ge2021yolox} for bootstrapping detections. 
Using an end-to-end benchmark allows us to evaluate improvements with the synthetic dataset in both detection and identification.

\noindent \textbf{Implementation Details.}
We follow the same hyperparameters and data preprocessing procedure from the author-provided MOTRv2 repository\footnote{\url{https://github.com/megvii-research/MOTRv2}} for all training jobs. 
For mixed training with our synthetic data, we simply combined both data from \texttt{{\datagenerator}RGB} and DanceTrack as one large dataset, without any additional probability sampling from the real and synthetic data.
All models were trained with 16 NVIDIA A4000 GPUs, using a batch size of 1. 
It took roughly 7 days of training for all synthetic and real data combined.

\subsection*{C.2 Group Activity Recognition}
Understanding collective human behaviors and social groups brings significant importance to various domains, including humanoid robots, autonomous vehicles, and human-computer interactions~\cite{chang2020transfer, cl2021acoustic, chang2022disentangling, chang2023importance, chang2022ivi, ehsanpour2022jrdb, martin2021jrdb, wu2021comprehensive}.
State-of-the-art methods for group activity recognition (GAR) leverage 2D skeletons as input due to the effectiveness and robustness gained from the less biased and more action-focused
representations~\cite{thilakarathne2022pose,zhou2022composer,perez2022skeleton,zappardino2021learning,zhai2022spatial,actortransformer}.
We describe the details of 2D skeleton-based GAR experiments that were primarily studied in our main paper. 
Let $[s_1,\cdots, s_t]$ denote a video with $t$ frames and $n$ persons, each frame $s_i:=\{p_1, \cdots, p_n\}$ where $p_i:=[(x_1, y_1, c_1), \cdots, (x_j, y_j, c_j)]$. Here, $j$ represents the number of joints in a person's skeleton, and each three-tuple $(x_j, y_j, c_j)$ respectively denotes the $x$ and $y$ coordinates in the pixel space and the class $c$ of the joint.
respectively denotes the $x$, $y$ coordinates in the pixel space and the class $c$ of the joint. 
For the given input $[s_1,\cdots, s_t]$, the objective of GAR is to output the class of the group activity performed by the dominant group among these $n$ persons and identify the action class of each individual in the video. 
Usually, the task assumes each video has one dominant group, as any outlier person does not contribute to the group.

\noindent \textbf{Implementation Details.}
Our implementations of Composer~\cite{zhou2022composer} and Actor Transformer~\cite{actortransformer} for the experiments are based on the open-sourced implementation and hyperparameter settings\footnote{\url{https://github.com/hongluzhou/composer}}.
The only modified hyperparameter is the batch size, from $384$ to $256$ due to our computation constraints with an NVIDIA RTX 3090 graphics card.
Both Composer and Actor Transformer are transformer~\cite{vaswani2017attention} based architectures.
Note that both architectures are slightly modified after synthetic pre-training, due to the differences between the synthetic and real datasets in the maximum number of persons in a clip and the number of atomic action classes.
Specifically, we set a different maximum sequence length to the transformer encoders of Composer and Actor Transformer and replaced the last layers (i.e., the group activity classifier and the person action classifier) with new classifiers to output the correct data shape for the target real-world dataset.

\subsection*{C.3 Controllable 3D Group Activity Generation}
Let $G^{TP} = \{m^{in}\}_{i=1 \sim T, n = 1\sim P}$ be a group of human motions with a total of $T$ frames and $P$ persons. 
The 3D pose of each person is denoted as $m^{in} \in R^{j\times d}$, where $j$ is the number of joints of a person and $d$ is the joint's feature dimension. 
We concatenate the global root position and all joints' 6D rotations~\cite{zhou2019continuity, chang2022ivi} as the pose representation.
Therefore, $j = 26$ (with $25$ actual joints) and $d = 6$.
The same representation is used as the input for 3D group activity recognition with Composer~\cite{zhou2022composer} and used as the ground truth during training for both MDM~\cite{tevet2022human} and MDM+IFormer baselines for 3D group activity generation. 

\begin{figure}[t]
\centering
   \includegraphics[width=0.78\linewidth]{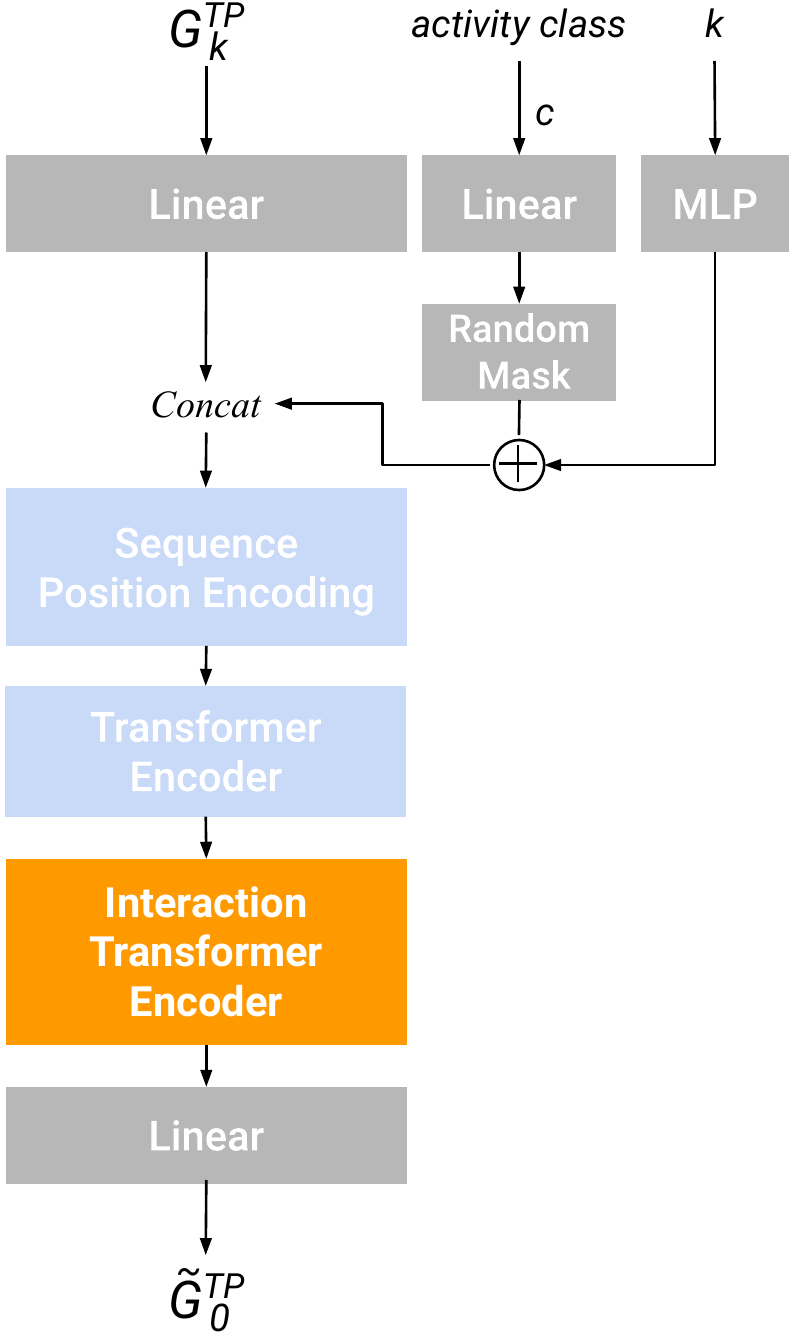}
   \caption{The architecture of MDM+IFormer. The model takes the
   noised
   group motion $G^{TP}_k$, an activity class, and a time step $k$ as inputs and outputs $\tilde{G}^{TP}_0$, an estimation of the clean group motion.
   The interaction transformer encoder is ``added'' after the transformer encoder layer of MDM~\cite{tevet2022human} for modeling the person interactions.}
   \vspace{-15pt}
\label{fig:mdmiformer_architecture}
\end{figure}

\noindent \textbf{Composer for 3D Group Activity Recognition.}
The learning-based metrics (recognition accuracy, FID, diversity, and multimodality)~\cite{guo2020action2motion} that are used to evaluate the generated 3D group activities require a well-trained 3D group activity \textit{recognition} model.
Composer~\cite{zhou2022composer} is a 2D skeleton-based group activity recognition model with a multi-scale transformer-based architecture. 
We chose Composer because it is a hierarchical architecture with fine-grained latent group-level and person-level features. 
We modified the first layer of Composer so that it accepts as input the aforementioned 3D group representation.
We used the same set of hyperparameters as the GAR experiments including a learning rate of $0.0005$, a weight decay of $0.001$, a hidden dimension of $256$ for the transformer encoders, except a batch size of $64$, a maximum number of $27$ persons, and a total number of $26$ joints. 

After the 3D group activity recognition model is well-trained, 
we obtain the 3D group activity recognition accuracy, and extract
latent group and person features to calculate the FID, diversity, and multi-modality metrics.
The latent group representation is the learned CLS token from the \textit{last} block and last scale of the Multi-Scale Transformer module of Composer, whereas the latent person representations are the learned person tokens from the \textit{last} block and the second scale (i.e., the transformer encoder of the person scale) of the Multi-Scale Transformer module.

\noindent \textbf{MDM \& MDM+IFormer.}
We follow the same diffusion scheme as MDM~\cite{tevet2022human} to obtain the noised group activity at every $k$-th diffusion time step, $G^{TP}_k$. 
Specifically, $G^{TP}_0 := G^{TP}$, meaning no noise is added at the $0$-th diffusion time step. 
The reversed diffusion process is then formulated as: 
\begin{equation}
    \tilde{G}^{TP}_0 = D(G^{TP}_k, k, c),
\end{equation}
where $D$ is the MDM+IFormer network, illustrated in  Fig.~\ref{fig:mdmiformer_architecture}. 
$\tilde{G}^{TP}_0$ is the estimated clean group activity and $c$ is the one-hot activity label.
The loss function follows the
objective in~\cite{ho2020denoising} and is defined as:
\begin{equation}
    L = \left|\left|G^{TP}_0 - \tilde{G}^{TP}_0\right|\right|^2.
\end{equation}
Our implementations of the MDM and MDM+IFormer baselines for 3D group activity generation are based on the author-released implementation of MDM\footnote{\url{https://github.com/GuyTevet/motion-diffusion-model}}
without any hyperparameter tuning.
Both models were optimized using the same loss function described above and trained on an NVIDIA RTX 3090 graphics card for 320K iterations.


\noindent \textbf{Formulas of Social Repulsive Forces}~\cite{helbing2000simulating}.
The three position-based metrics (in \red{Sec. 4.3.1} of the main paper) are formulated as follows.

\noindent \textit{-- Repulsive interaction force:} 
\begin{equation} \label{eq:interaction_force}
    \vec{f}_{ij}^{int} = A \cdot exp[(r_{i} + r_{j} - d_{ij})/B] \cdot \vec{n}_{ij}.
\end{equation}
$\vec{f}_{ij}^{int}$ is the interaction force of character $j$ applied to character $i$. $A$ and $B$ are constants ($A:=2,000$ and $B:=0.08$). $r_{i}$ and $r_{j}$ are the radius of the characters. $d_{ij}$ is the distance between the characters. $\vec{n}_{ij}$ is the unit vector pointing from character $j$ to character $i$.

\noindent \textit{-- Contact repulsive force:}
\begin{equation} \label{eq:interaction_force}
    \vec{f}_{ij}^{cont} = k \cdot max(0, r_{i}+r_{j}-d_{ij}) \cdot \vec{n}_{ij},
\end{equation}
where $k$ is a constant ($k:=120,000$). 

\noindent \textit{-- Total repulsive force:}
\begin{equation} \label{eq:interaction_force}
    \vec{f}_{ij}^{total} = \vec{f}_{ij}^{int} + \vec{f}_{ij}^{cont}.
\end{equation}
All constants follow the social force model proposed in~\cite{helbing2000simulating}.

\noindent  \textbf{Generated 3D Group Activities.} Please refer to our supplementary video for the rendered 3D group activities generated by MDM and MDM+IFormer.



\section*{D. Additional Experiments}

\begin{table}[]
\small
  \aboverulesep=0ex
  \belowrulesep=0ex 
\begin{center}

\begin{tabular}{l|l|c}
\toprule
\multicolumn{2}{l|}{\textbf{\#Epochs required for model convergence}} & \textbf{Target}   \\ \cline{3-3}
\multicolumn{2}{c|}{(Composer~\cite{zhou2022composer} / Actor Transformer~\cite{actortransformer})}   & CAD2~\cite{choi2011learning}     \\ \midrule
\multirow{2}{*}{\textbf{Source}}                     & CAD2~\cite{choi2011learning}                     & 88 / 233 \\
                                            & \texttt{\datagenerator}                   & 13 / 92  \\ \bottomrule
\end{tabular}
\end{center}
\vspace{-10pt}
\caption{Pre-training with our synthetic data leads to faster model convergence on the target domain for GAR. (Composer: $6.8\times$ faster; Actor Transformer: $2.5\times$ faster) }
\label{tab:convergence}
\end{table}

Pretraining with data from {\datagenerator} can improve convergence speed on the target dataset. We conduct the GAR experiment using 2D skeletons as the only input modality for both models and compare the number of epochs required for model convergence in Tab.~\ref{tab:convergence}. 
To automatically determine whether or not the model training has saturated, we adopted early stopping by setting a maximum number of $500$ epochs with stopping patience of $50$ epochs. 
The results suggest that training Composer from scratch on CAD2 requires 88 epochs on average; with \texttt{{\datagenerator}} pre-training, Composer only requires 13 epochs for convergence.

\section*{E. Limitations \& Future Work}
We demonstrate that synthetic data can replace a great amount of real data \cite{chang2024equivalency} 
and successfully mitigate the scarcity of real data for multi-person and multi-group tasks, despite the domain gap that restricts the generalizability of models trained with synthetic data. 
With the release of our data generator, {\datagenerator}, we encourage the community to create their own data or enhance the synthetic data generator.
While collecting more assets and generating more data with adjusted camera views can surely increase data diversity and shorten the gap,
we would also like to point out some aspects of the generator that should be addressed in the future to create more realistic data.

\noindent \textbf{Publicly-available assets.}
Most assets we use in {\datagenerator} are publicly available, including HDRIs, human characters, and animations. 
However, most existing assets such as photometric 3D scenes and high-quality avatars may require additional licensing for model training. 
Some assets could be restricted to specific game engines, which hinders the development of synthetic data. 

\noindent \textbf{Simulated Hair and clothes.}
The avatars used in {\datagenerator} do not contain hair and clothes physics that are in accordance with their body motions. 
Adding cloth and strand-based hair simulation would be sufficient for realistic interactions between hair/cloth and body, and thus improve the data quality.

\noindent \textbf{Finger and face Movements.}
Our animations do not contain finger and face movements. 
While it might be reasonable as human groups are generally captured from a distance, adding finger and face movements can still improve the fidelity of the human motions.

\noindent \textbf{Human-environment interactions.}
Like most synthetic datasets \cite{black2023bedlam, yang2023synbody, varol2017learning, bazavan2021hspace}, {\datagenerator} lacks meaningful interactions between human and environment. 
The interactions might include groups of humans navigating in a complex environment, a person picking up a phone while texting, or holding a suitcase.
Animating human motions and activities with scene awareness is incredibly challenging. 
A simple solution is to polish each scene by carefully placing the avatars and staging the human behaviors. However, it would require significant manual efforts and limit the scale and diversity of the synthetic data.

\noindent \textbf{Complexity for human groups.}
Animating human groups is significantly more challenging than animating the motions of a single person because the complexity (the number of interactions) increases quadratically as the number of persons increases.
We apply relatively simple heuristics when designing rules for authoring human groups, which could only reflect a certain portion of real-world activities. 
These underlying rules that drive the group motions could also lead to datasets that are less complex than real-world ones, limiting the model generalization on downstream tasks.
Creating new groups with expert-guided heuristics, LLM-generated rules, or directly from the 3D GAG method, should be considered.
An alternative would be using existing motion capture data in replace of the procedural generation method. 
However, the lack of fine-grained motion capture data for large-scale collective 3D group motions is an obstacle to the development.

\noindent \textbf{Societal Impacts.}
While we demonstrate the effectiveness of our synthetic data on several tasks, it is important to note that the use of synthetic data, in all manners, may still result in unbalanced and biased results.
We strive to ensure the inclusiveness and fairness of our datasets by incorporating human avatars of all ages, genders, and ethnicities, providing a representative and equitable approach to generating data for responsible advancement in related fields.

\end{document}